%% file: main.tex
\documentclass[10pt,twocolumn,letterpaper]{article}

\usepackage[pagenumbers]{cvpr} % To force page numbers, e.g. for an arXiv version

\input{preamble}
\definecolor{cvprblue}{rgb}{0.21,0.49,0.74}
\usepackage[pagebackref,breaklinks,colorlinks,allcolors=cvprblue]{hyperref}

\usepackage{xcolor} 
\usepackage{color, colortbl}
\usepackage{arydshln}
\usepackage{booktabs} 
\usepackage{pifont}
\usepackage{tcolorbox}
\usepackage{fontawesome}

\newcommand{\dashedmidrule}{\noalign{\vskip\aboverulesep}\hdashline\noalign{\vskip\belowrulesep}}

\def\paperID{\#} % *** Enter the Paper ID here #20282
\def\confName{CVPR}
\def\confYear{2026}

\title{CoVR-R: Reason-Aware Composed Video Retrieval}

\author{\\ {Omkar Thawakar}\textsuperscript{1$\dagger$} \quad{Dmitry Demidov}\textsuperscript{1$\dagger$} \quad{Vaishnav Potlapalli}\textsuperscript{$\dagger$}\quad {Sai Prasanna Teja Reddy Bogireddy}\textsuperscript{2$\dagger$} \\
    {Viswanatha Reddy Gajjala}\textsuperscript{3$\dagger$} \quad  {Alaa Mostafa Lasheen}\textsuperscript{1} \quad  
     {Rao Muhammad Anwer}\textsuperscript{1} \quad  {Fahad Khan}\textsuperscript{1,4}\\
     \fontsize{12pt}{12pt}\selectfont \textsuperscript{1}Mohamed bin Zayed University of AI,  \textsuperscript{2}University of Chicago, \\
     \textsuperscript{3}University of Wisconsin-Madison,
     \textsuperscript{4}Linköping University \\
     \fontsize{10pt}{12pt}\selectfont \faEnvelopeO \hspace{2pt} \{{omkar.thawakar, dmitry.demidov}\}@mbzuai.ac.ae \\
 {\hypersetup{urlcolor=blue}
\fontsize{11pt}{12pt}\selectfont \faExternalLink \hspace{2pt} \href{https://mbzuai-oryx.github.io/CoVR-R/}{https://mbzuai-oryx.github.io/CoVR-R/}}
}

\begin{document}
\maketitle
\input{sec/0_abstract}   
\def\thefootnote{$\dagger$}\footnotetext{Equal contribution.}
\input{sec/1_intro}

\input{sec/2_related_work}

\input{sec/3_benchmark}

\input{sec/3_Method}
\input{sec/4_Experiments}

\input{sec/5_Conclusion}
{
    \small
   \bibliographystyle{ieeenat_fullname}
   \bibliography{main}
}

\input{sec/X_suppl}
% {
%     \small
%     \bibliographystyle{ieeenat_fullname}
%     \bibliography{main}
% }

\end{document}

%% file: sec/0_abstract.tex
\begin{abstract}
Composed video retrieval (CoVR) aims to find a target video given a reference video and a textual modification. Prior work assumes the modification text fully specifies the visual changes, overlooking after-effects implicit consequences (e.g., motion, state transitions, viewpoint/duration cues) that emerge from the edit. We argue that successful CoVR requires reasoning about these after-effects. We introduce a reasoning-first, zero-shot approach that leverages large multimodal models to (i) infer causal and temporal consequences implied by the edit, and (ii) align the resulting reasoned queries to candidate videos without task-specific finetuning. To evaluate reasoning in CoVR, we also propose CoVR-Reason, a benchmark that pairs each (reference, edit, target) triplet with structured internal reasoning traces and challenging distractors that require predicting after-effects rather than keyword matching. Experiments show that our zero-shot method outperforms strong retrieval baselines on recall@K and particularly excels on “implicit-effect” subsets. Our automatic and human analysis confirm higher step-consistency and effect-factuality in our retrieved results. Our findings suggest that general-purpose LMM reasoning is an effective driver for CoVR, reducing the need for task-specific supervision and opening a path toward more explainable video search. Our model, code and benchmark is available at: \url{https://github.com/mbzuai-oryx/CoVR-R}. 
\end{abstract}

%% file: sec/1_intro.tex
\section{Introduction}
\label{sec:intro}

In composed video retrieval (CoVR), a system receives a reference video and a short modification text and must return a target video that reflects the requested change. Real queries, however, are rarely fully explicit. In practice, many modifications imply additional, unspoken after-effects causal and temporal consequences such as object state transitions (“ingredients become browned”), motion and shot-scale changes (“close-up implies tighter framing and shorter duration”)\cite{cinescale2021}, or scene dynamics (“frying introduces smoke and faster hand motions”). Treating these after-effects as mere keywords underestimates the gap between what is said (the edit) and what must happen (its visual consequences). We argue that closing this gap requires reasoning \emph{i.e.} predicting effect chains that connect the edit to plausible video evidence. 
% The edit “same scene but now the chef starts frying; switch to a close-up” implies consequences (oil shimmer, faster hand motion, tighter framing, louder sizzle) that are not literally stated. Likewise, “keep the player, but move to the post-goal celebration in the stands” hints at a temporal jump, new crowd dynamics, and different shot scale/background, even if those words never appear in the edit. These after-effects—causal and temporal consequences of an edit—are central to user intent but are typically unmodeled in current systems.

\begin{figure*}[ht!]
\includegraphics[width=\textwidth]{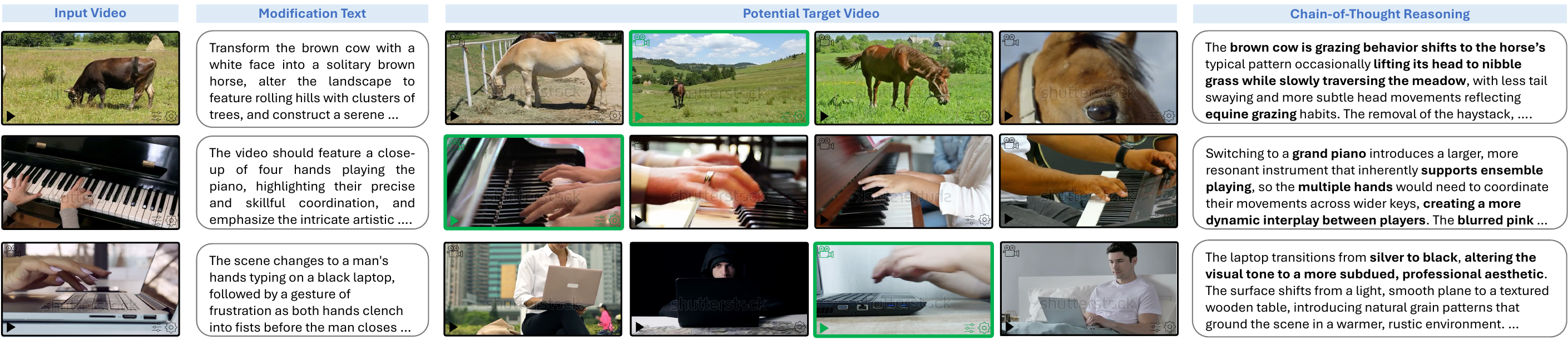}
% \vspace{-2em}
\caption{
\textbf{Why reasoning is needed for CoVR}. Each row pairs a reference video and edit text with the desired target (highlighted in green), where success depends on after-effects rather than keyword overlap. Examples include: shifting from a brown cow to a solitary brown horse in rolling hills with trees (state and scene changes); switching to a close-up of four hands on a piano (cinematography and object configuration); moving from typing to clenched fists and closing the laptop (temporal phase progression). These cases require inferring consequences such as state transitions, phase order, and shot scale that not merely matching words. Reasoning bridges what the edit says and what the target must show, capturing the visual consequences that define correct retrieval. More examples are in suppl. material.
}
\label{fig:intro_figure}
% \vspace{-1em}
\end{figure*} 

Early progress on composed image retrieval (CoIR) established the paradigm of combining a reference visual input with a text edit to find a target, along with benchmarks such as FashionIQ \cite{guo2019fashion} (fine-grained fashion edits) and CIRR \cite{cirr} (open-domain, human-written modifications). Representative efforts include early feature-modulation and local-feature methods \cite{vo2019tirg}, and the move to open-domain datasets such as CIRR \cite{radford2021clip, li2022blip, li2023blip2}, which highlighted ambiguity and the need to identify the object of interest amid distractors. Surveys now catalog modeling choices \cite{song2025cirsurvey} (feature fusion, text-conditioned transforms, training objectives), yet the majority of CIR settings remain edit-literal rather than effect-reasoned. The image literature thus frames the task but does not fully confront temporal/causal consequences that arise in video.

% CoVR extends CoIR into the temporal domain. Recent methods introduce stronger video encoders \cite{webvid-covr}, enriched textual context, and discriminative embeddings \cite{thawakar2024composed}, and even propose automatic data construction \cite{ventura2024covr} to scale training. These works improve recall by better mixing visual and textual cues or by enlarging supervision, but they typically remain triplet-driven (reference, edit, target) and principally reward keyword-compatible matches. As a result, they can fail on edits whose success depends on implied consequences rather than surface overlap (e.g., a camera move or phase transition that the text hints at but does not spell out). Figure \ref{fig:intro_figure} illustrates representative cases where retrieval success hinges on inferring such after-effects: transitioning from a grazing cow to a horse in rolling hills requires anticipating changes in gait and grazing behavior; switching to a four-hand piano performance necessitates reasoning about spatial coordination and instrument scale; and progressing from typing to clenched fists demands understanding temporal phase sequences and gestural transitions. Our work takes a complementary path: instead of learning a task-specific edit combiner, we explicitly reason about after-effects and use those structured inferences to guide retrieval in a zero-shot manner without CoVR-specific finetuning.
CoVR extends CoIR into the temporal domain. Recent works strengthen video encoders \cite{webvid-covr, bain2021frozen}, enrich textual context and embeddings \cite{thawakar2024composed}, and scale training via automatic data construction \cite{ventura2024covr}. While these advances improve recall, most remain triplet-driven (reference–edit–target) and primarily reward keyword overlap \cite{pic2word, circo}. Consequently, they miss edits whose success depends on \emph{implied} consequences e.g., a camera move or phase transition only hinted at by the text. As shown in Figure~\ref{fig:intro_figure}, success often hinges on reasoning about after-effects: cow→horse with terrain cues, four-hand piano close-ups, or typing→fist-clench→close sequences. These limitations motivate a shift from triplet matching to consequence modeling. Inspired by this, in this work we explicitly infer the edit’s after-effects (state, phase, scene, camera, tempo) with strong multimodal model and use them to drive retrieval, yielding a zero-shot CoVR pipeline that requires no task-specific fine-tuning.
% cite and mention our ICCV paper in above paragraph

Large multimodal models (LMMs) have recently demonstrated emergent reasoning across perception \cite{Yang2025R1OnevisionAGA,Zhang2025R1VLLTA,Wang2025VisualPRMAEA,Peng2025SkyworkRPA,thawakar2025llamav}, language, and video understanding \cite{wang2025internvideo2,schneider2025quickvideo,yuan2025date}, including long-context temporal comprehension. In particular, Qwen3-VL \cite{yang2025qwen3} reports strong visual reasoning and video dynamics understanding, making it a natural candidate to act as a reasoner in retrieval pipelines. We leverage these capabilities to (i) infer compact, schema-constrained after-effect summaries conditioned on the reference video and edit (e.g., state changes, action phases, camera/shot cues, tempo), and (ii) transform those summaries into effect-aware queries that re-rank candidate videos. Crucially, our framework keeps the LMM backbone frozen and does not rely on CoVR-specific supervision; the LMM is used for reasoning and checking, not for learning a new retriever from scratch.

To fairly assess reasoning, we introduce a benchmark that pairs each triplet with \emph{reasoning traces} and \emph{hard distractors} to defeat keyword-only matching. Unlike prior CoVR resources \cite{webvid-covr,thawakar2025beyond} that emphasize literal edits or caption alignment, our benchmark stresses \emph{causal plausibility} and \emph{temporal consistency} (e.g., sub-action order, expected shot scale, correct state change), and adds diagnostics for \emph{step-consistency} and \emph{effect-factuality} alongside standard retrieval metrics. This reason-supervised evaluation aligns with recent trends toward richer supervision. Methodologically, we take a route orthogonal to training-heavy edit-fusion pipelines: we import LMM reasoning to predict effects and verify them, exposing an explicit chain used for both query expansion and re-ranking. Finally, unlike CoIR settings where training-free behavior sometimes suffices, we directly target video-specific after-effects—temporal order, shot scale, and action phases and provide a benchmark that evaluates them.

\noindent Our contributions can be summarized as follows, 
\begin{itemize}
    \item We introduce a reasoning-aware CoVR benchmark (\emph{CoVR-R}) with internal reasoning targets and hard distractors, enabling evaluation that goes beyond keyword matching.
    \item We propose reason-aware composed video retrieval framework using Qwen3-VL-8B that explicitly predicts after-effects implied by the edit and uses them to guide retrieval in zero shot setting. Concretely, we prompt Qwen3-VL to produce a compact, structured reasoning record (states, actions, scene, camera, tempo) conditioned on the reference video and edit, transform it into effect-aware queries.
    \item We show consistent gains of $16\%$ compared to previous methods on our proposed \emph{CoVR-R} benchmark that requires implicit effects where mere word overlap is insufficient. Our method also exceeding strong baselines on Dense-WebVid-CoVR by the margin of $12\%$. Together, these results suggest that general-purpose LMM reasoning is an effective driver for CoVR, reducing dependence on task-specific annotation and bringing retrieval behavior similar to users describe edits in the wild.
\end{itemize}

%% file: sec/2_related_work.tex
\section{Relation to Prior Arts}
\label{sec:related_work}

\noindent\textbf{Composed Image Retrieval (CoIR).}
Composed image retrieval formulates search as conditioning on a \emph{reference image} plus a \emph{edit text}, returning an image that reflects the specified change \cite{vo2019composing}. 
Early progress was driven by carefully curated, human-annotated resources such as CIRR~\cite{cirr} and FashionIQ~\cite{guo2019fashion}. 
While these datasets offer high-quality supervision, their scale is naturally capped by annotation cost. 
To overcome this bottleneck, recent work explores automatic data construction at scale, e.g., LaSCo~\cite{levy2023data} and SynthTriplets18M~\cite{gu2023compodiff}, often leveraging VQA pipelines and text-guided editing systems like InstructPix2Pix~\cite{brooks2022instructpix2pix} to synthesize modification triples. 
% However, several of these large-scale resources are not yet publicly accessible, limiting reproducibility and broad evaluation.

\noindent\textbf{Composed Video Retrieval (CoVR).}
In CoVR, the goal is to retrieve a \emph{target video} given a \emph{reference video} and a \emph{textual edit}. A common strategy is to adapt CoIR techniques to the temporal setting by aggregating multi-frame or clip-level features and aligning them with the edit text~\cite{rasheed2023fine,xu2021videoclip,xue2022clip,yang2021taco}. The availability of large web video corpora and vision–language models (VLMs)~\cite{li2023blip,radford2021learning} has further advanced CoVR: for instance, WebVid-CoVR~\cite{webvid-covr} mines caption-similar video pairs and uses LLMs to produce modification prompts for contrastive training. Beyond this, \cite{thawakar2024composed} improves query–video alignment by incorporating language descriptions of the source video
%, and \cite{hummel2024egocvr} proposes an egocentric action-retrieval benchmark focused on first-person activities. The latter is valuable yet limited to a test-only release and a specific viewpoint/domain, lacking dense edit text and the breadth of general-purpose content (e.g., nature, sports, educational, lifestyle) present in WebVid-CoVR~\cite{webvid-covr}.
Complementary to these efforts, \emph{Dense-WebVid-CoVR} enriches the composed retrieval setting with more detailed edit descriptions and harder negatives that emphasize multi-step, temporally grounded changes, thereby stressing reasoning beyond surface keyword overlap~\cite{thawakar2025beyond}.

\begin{figure*}[t]
\includegraphics[width=\textwidth]{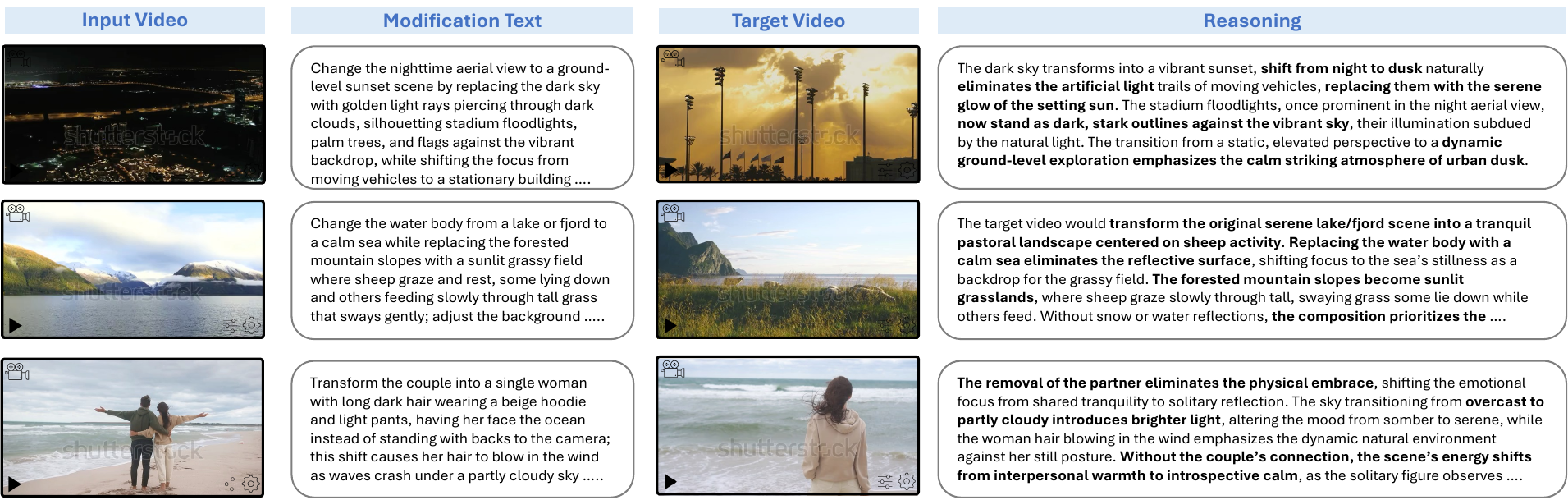}
% \vspace{-2em}
\caption{
\textbf{Data samples from our CoVR-R benchmark}. Each row shows an input video, its edit text, the target video, and a concise reasoning trace that makes the implied after-effects explicit (e.g., state/scene shifts, camera/shot changes, temporal phases). Examples illustrate: nighttime aerial → ground-level sunset with silhouetted floodlights and palm trees; lake/fjord → calm sea with sunlit grassy field and sheep activity; couple → solitary woman facing the ocean with wind-blown hair under partly cloudy sky. These annotations ground what the edit implies (not just what it says) and guide evaluation of reasoning-aware retrieval. % More examples are in suppl. material.
}
\label{fig:architecture}
% \vspace{-1em}
\end{figure*} 

\noindent\textbf{Reasoning in Retrieval.}
Recent work in composed image retrieval (CoIR) increasingly treats reasoning over history, semantics, and implicitly address as an important capability rather than a byproduct of embedding fusion.
There are several complementary threads have emerged in a reasoning for retrieval landscape. \citet{chen2025mai} proposes a multi–turn CIR framework 
%(Two–stage Semantic Aggregation + Multi–turn Iterative Optimization) 
that explicitly aggregates historical turns to avoid “last–turn shortcuts,” yielding stronger multi–turn reasoning and alignment. \citet{tang2025reason} introduces a training–free, one–stage reflective chain–of–thought that reasons over the manipulation text with contextual cues from the reference image.
%, then retrieves in a single pass.% 
ZS-CIR \cite{tu2025multimodal} eliminates fragile intermediate target texts by using an MLLM “reasoning agent” to construct high–quality triplets directly from unlabeled images, enabling zero–shot CIR with sizable gains. MVFT-JI \cite{tu2025mllm} jointly leverages an MLLM to generate target texts and to reason at inference, aligning composed queries and candidates without labeled triplets and improving zero–shot CoIR. Collectively, these directions move beyond keyword–compatibility to predict \emph{consequences} of an edit an outlook we generalize to video.
While these works demonstrate the value of explicit or implicit reasoning in CoIR, they operate on static images; extending the same principles to composed \emph{video} retrieval (CoVR) requires predicting \emph{after–effects} that are inherently \emph{temporal} (phase order, duration/tempo) and \emph{cinematographic} (shot scale, camera motion), and then verifying those effects in candidate videos.

\paragraph{Why reasoning-based CoVR ?}
Across both CoIR and CoVR, much of the supervision remains \emph{triplet-centric} and often \emph{edit-literal}: models are rewarded for matching explicit phrases rather than for anticipating the \emph{after-effects} an edit implies (e.g., state transitions, temporal ordering, shot-scale or camera changes). 
Large-scale synthetic triplets~\cite{levy2023data,gu2023compodiff} and mined web pairs~\cite{webvid-covr} grow coverage, but they do not expose or evaluate an internal chain of reasoning. 
Dense-WebVid-CoVR~\cite{thawakar2025beyond} takes a step toward denser, more informative edits but still lacks explicity reasoning requirements.
%the requirement for reasoning for compositional retrieval.
Our goal is to explicitly brings \emph{reasoning} into the retrieval loop:
%for more precise retrieval in challenging scenes:
given a reference video and an edit text, we predict the likely consequences (the “after-effects”) and use them to drive target retrieval. This reasoning-first perspective aims to bridge the gap between what the edit \emph{states} and what the target video must \emph{show}, yielding more faithful and robust composed video retrieval.
Our approach achieves this by (i) using a strong LMM to infer structured, schema–constrained after–effects conditioned on (reference, edit), (ii) converting them into reason–aware queries for coarse retrieval, and (iii) checking them with a lightweight, slot–wise verifier thereby bringing the “reason–then–retrieve’’ paradigm to CoVR where consequences, not just keywords, determine success. To faithfully measure this capability, we also build \textsc{CoVR-R}, a reason–aware CoVR benchmark composed of challenging, real–world–like scenarios with dense reasoning traces and hard distractors, which both evaluates our proposed method and stress–tests competing systems on temporal, causal, and cinematographic after–effects.

%% file: sec/3_benchmark.tex
\section{The \textsc{CoVR-R} Benchmark}
\label{sec:covrr_benchmark}

Our proposed \textsc{CoVR-R} (Composed Video Retrieval with \emph{Reasoning}) is a curated benchmark that emphasizes \emph{after–effect} reasoning in composed video retrieval.
Unlike prior triplet sets that are largely \emph{edit–literal}, \textsc{CoVR-R} pairs each triplet with grounded, schema–constrained reasoning traces and hard distractors, and it focuses on cases where success depends on causal/temporal and cinematographic consequences of the edit.
Our benchmark contains {\emph{2800}} high–quality triplets with verified reasoning annotations. A visual overview with examples are presented in Figure~\ref{fig:intro_figure}.

\subsection{Source corpora and triplet construction:}
We build \textsc{CoVR-R} from two complementary sources: \emph{Dense–WebVid–CoVR}~\cite{thawakar2025beyond} (open–domain diversity with dense descriptions) and \emph{Something–Something V2 (SSv2)} \cite{goyal2017something} with short, object–centric actions with clear temporal phases.
From Dense–WebVid–CoVR, we begin with candidate composed triplets and retain only those that \emph{require} reasoning.
A candidate is accepted if it satisfies at least two of the following: (i) temporal dependency (the edit implies an ordered phase change); (ii) state transition (a visible change such as raw$\!\rightarrow\!$ browned or empty$\!\rightarrow\!$ filled); (iii) cinematography (implied camera/shot consequences such as tighter shot, pan, or zoom); (iv) implicit cause–effect (the edit hints outcomes not literally stated, e.g., rain $\Rightarrow$ wet surfaces); and (v) low lexical sufficiency (keyword overlap between $E$ and $V_t$’s description is insufficient to retrieve $V_t$ without reasoning).
For SSv2, which does not natively provide composed triplets, we synthesize reasoning–dependent pairs by (a) pairing clips that share objects but differ by \emph{phase} or \emph{state} outcome; (b) generating a concise edit $E$ that implies the change (e.g., ``now the liquid is poured and the cup is full''); (c) keeping only pairs where the difference cannot be resolved by a single keyword; and (d) mining hard negatives that match objects/verbs but violate phase order or shot scale.
This procedure yields compact edits whose satisfaction demonstrably requires temporal/state and cinematographic reasoning.

\subsection{Reasoning Trait Generation}
For each curated triplet $(V_r,E,V_t)$ we create a reasoning record such that:
\vspace{-0.5em}
\begin{equation}
\small
    R=\{\texttt{states},\ \texttt{actions},\ \texttt{scene},\ \texttt{camera},\ \texttt{tempo}\}
\end{equation}
% \vspace{0.5em}

% where each key contains at most four \emph{atomic} assertions expressed in relative, verifiable terms (e.g., \texttt{states: browned–surface}, \texttt{actions: celebration–in–stands}, \texttt{camera: tighter–than–reference}, \texttt{tempo: short–cut}).
For each slot in $R$, we allow \emph{at most four} \emph{atomic} assertions. 
An assertion is \emph{atomic} if it (i) names a single, observable effect; (ii) uses one controlled predicate from the slot vocabulary; and (iii) can be checked over a concrete time span of $V_t$ \emph{relative} to $V_r$ (e.g., “tighter than in $V_r$”). 
We ban compounding (“and/then”) and adjectives that are not tied to a measurable cue. 
Each assertion is written in a canonical \texttt{slot: value} form, optionally with a relative operator (\texttt{>} / \texttt{<} / \texttt{=} to $V_r$) and an evidence tag (time range or frame index).
Reasonig traits are produced using a constrained prompt that references the dense descriptions of $V_r$ and $V_t$ together with the edit $E$, followed by a canonicalization pass that (i) removes duplicates, (ii) resolves contradictions by keeping the higher confidence assertion and demoting the other to \emph{verify–only}, and (iii) emits assertions in a fixed slot order (actions $\rightarrow$ camera $\rightarrow$ states $\rightarrow$ scene $\rightarrow$ tempo) to stabilize downstream text encoding.
The final text form of $R$ (``effect query'' $Q_R$) and a binary checklist derived from $R$ are stored with each triplet.

\begin{figure*}[t]
\includegraphics[width=\textwidth]{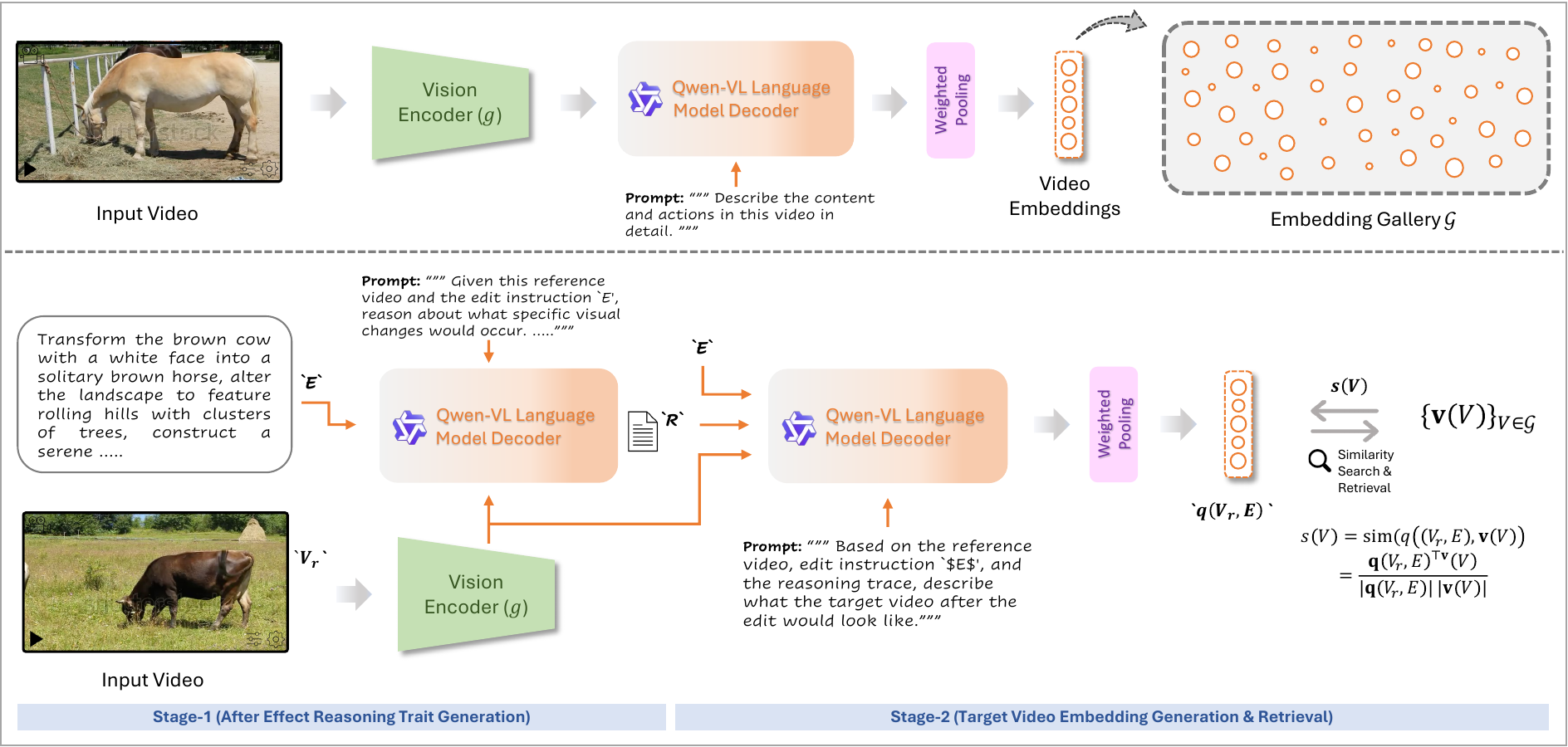}
% \vspace{-1.5em}
\caption{
\textbf{Overview of our \emph{Reason–then–Retrieve} architecture}. Given reference video $V_r$ and edit text $E$, Stage-1 uses Qwen3-VL-8B to generate an \emph{after-effect reasoning trace} $R$ predicting concrete consequences (actions/phases, camera/shot, states, scene, tempo) conditioned on $V_r$ and $E$. \emph{Stage–2} converts $(V_r,E,R)$ into an effect-aware query embedding $\mathbf{q}(V_r,E,R)$ and performs cosine retrieval against precomputed gallery embeddings ${\mathbf{v}(V)}_{V\in\mathcal{G}}$ %from a frozen vision encoder 
via weighted temporal pooling. This reasoning-first pipeline delivers zero-shot CoVR by turning edits into explicit consequences for matching. 
}
\label{fig:architecture}
% \vspace{-1em}
\end{figure*} 

\subsection{Grounding and Manual Correction}
\label{sec:human_eval}
Each assertion in $R$ is \emph{grounded} against both videos under three rules: (1) there must exist at least one frame or time range in $V_t$ that visibly supports the assertion; (2) the same assertion must be absent or strictly weaker in $V_r$ (else it is rewritten in explicit relative form or discarded); and (3) the phrasing must be unambiguous and template consistent.
We apply a two–stage human review: a senior pass validates slot coverage and removes stylistic language; a cross–audit swaps batches across annotators to detect leakage (e.g., identity cues embedded in scene descriptions).
Disagreements are resolved by evidence first voting; inter–annotator agreement (Cohen’s $\kappa$) is reported in the suppl. material.
Where automatic drafts are low–confidence or contradictory, we either reduce them to \emph{verify–only} checks or remove them if visual evidence is insufficient.

%% file: sec/3_Method.tex
\section{Method}

\noindent\textbf{Task:}
Given a reference video $V_r$ and a modification text (``edit'') $E$, the goal is to retrieve a target video $V_t$ from a gallery $\mathcal{G}$ that best reflects the \emph{after-effects} implied by $(V_r,E)$. Unlike prior work that treats $E$ as literal keyword constraints, we explicitly reason about visual consequences of state transitions, temporal phases, and cinematographic changes to retrieve via semantic alignment in a zero-shot manner.

\subsection{Overall Architecture}
\label{subsec:overall}
% Our reasoning-first pipeline operates in two stages (Fig.~\ref{fig:teaser}):
A simple way to build a zero-shot, reasoning-aware CoVR system is to start from existing CoIR pipelines and extend them to videos. We therefore adapt MVFT-JI to the video setting (denoted MVFT-JI$^\dagger$), keeping its cross-attention joint inference over frame features and the original text-fusion design. However, this adaptation offers only modest gains: the reasoning remains mostly surface-level (keyword alignment) and struggles with effects crucial in CoVR such as temporal phase progressions (before→after), cinematography changes (shot scale, pan/zoom), and subtle state transitions. Motivated by these limits, our overall method built on top of a strong multimodal reasoner (Qwen-VL) to infer a structured after-effect trace from $(V_r, E)$ convert it into an effect-aware query for coarse retrieval.
Our approach operates in two stages (Figure.~\ref{fig:architecture}): \textbf{(1)} reasoning-based query generation from $(V_r, E)$, and \textbf{(2)} semantic matching against pre-encoded gallery representations. We detail the gallery encoding below, with query generation described in Sec.~\ref{subsec:query}.

\subsection{Gallery Video Encoding}
\label{subsec:gallery}

% For each gallery video $V\in\mathcal{G}$, we construct a semantic embedding as follows:

For each $V\in\mathcal{G}$, we construct a semantic embedding $\mathbf{v}(V)\in\mathbb{R}^d$ via vision-language grounding and importance-weighted aggregation.

\noindent\textbf{Step 1: Video description generation.}
We feed $V$ to Qwen3-VL with the prompt: 
\vspace{-0.5em}
\begin{quote}
\emph{``Describe the content and actions in this video in detail.''}
\end{quote}
\vspace{-0.5em}
The model generates a natural language description $D(V)$ consisting of $N$ tokens: $\{w_1,w_2,\ldots,w_N\}$.

\noindent\textbf{Step 2: Extract token embeddings.}
We extract the final output layer embeddings before the decoding head.
For each generated token $w_i$, we obtain its corresponding embedding $\mathbf{h}_i\in\mathbb{R}^d$ from the last transformer layer, yielding a sequence $\{\mathbf{h}_i\}_{i=1}^N$ with shape $(N,d)$.

\noindent\textbf{Step 3: Importance Weighted pooling.}
To aggregate the token embeddings into a single video representation, we apply weighted mean pooling:
\vspace{-.5em}
\begin{equation}
\label{eq:weighted_pool}
\mathbf{v}(V)=\frac{\sum_{i=1}^N \alpha_i \mathbf{h}_i}{\sum_{i=1}^N \alpha_i},
\end{equation}
where $\alpha_i$ is the importance weight for token $w_i$.
We assign higher weights to semantically rich tokens (action verbs, object nouns, descriptive adjectives) and lower weights to generic function words (articles, prepositions, auxiliary verbs).
Specifically, we use a lexical category-based weighting scheme:

\vspace{-0.5em}
\begin{equation}
\label{eq:weighted_pool_alphas}
\alpha_i = \begin{cases}
\alpha_{\text{high}} & w_i \in \mathcal{C}_{\text{salient}} \text{ (actions, objects, states)} \\
\alpha_{\text{mid}} & w_i \in \mathcal{C}_{\text{context}} \text{ (attributes, scene)} \\
\alpha_{\text{low}} & w_i \in \mathcal{C}_{\text{generic}} \text{ (function words)}
\end{cases}
\end{equation}

% \begin{equation}
% \alpha_i = \begin{cases}
% 1.0 & \text{if } w_i \text{ is action/object/state descriptor} \\
% 0.5 & \text{if } w_i \text{ is modifier or scene element} \\
% 0.1 & \text{if } w_i \text{ is generic/functional word}
% \end{cases}
% \end{equation}

\noindent Weights are determined by mapping tokens to lexical categories via the model's tokenizer and a predefined category dictionary. All gallery embeddings $\{\mathbf{v}(V)\}_{V\in\mathcal{G}}$ are precomputed \emph{offline} and cached for retrieval.

%------------------------------------------------------------------------
\subsection{Query Encoding with Edit Reasoning}
\label{subsec:query}

Given a reference video $V_r$ and edit instruction $E$, we generate a representation of the desired target video through a two-step reasoning process:

\noindent\textbf{Step 1: After-effect reasoning.}
We prompt Qwen3-VL with $(V_r,E)$ to produce a structured reasoning trace $R$ that explicitly identifies the visual consequences of applying the edit.
The prompt is:
\begin{quote}
\emph{``Given this reference video and the edit instruction `$E$', reason about what specific visual changes would occur. List the expected changes in: (1) object states, (2) actions or phases, (3) scene or background, (4) camera or framing, (5) tempo or pacing. Provide a structured reasoning trace.''}
\end{quote}
The model outputs $R$, a structured record enumerating expected transformations across these dimensions.

\noindent\textbf{Step 2: Target description generation.}
We then condition Qwen3-VL on $(V_r, E, R)$ to generate a complete description of the hypothetical post-edit video:
\begin{quote}
\emph{``Based on the reference video, edit instruction `$E$', and the reasoning trace, describe what the target video after the edit would look like.''}
\end{quote}
This produces a target description $D_{\text{target}}$ consisting of $M$ tokens $\{w_1',w_2',\ldots,w_M'\}$.

\noindent\textbf{Step 3: Extract and pool embeddings.}
Following the same procedure as in Sec.~\ref{subsec:gallery}, we extract the final layer embeddings $\{\mathbf{h}_i'\}_{i=1}^M$ for the generated target description and apply weighted pooling:
\begin{equation}
\mathbf{q}(V_r,E)=\frac{\sum_{i=1}^M \alpha_i' \mathbf{h}_i'}{\sum_{i=1}^M \alpha_i'},
\end{equation}
where weights $\alpha_i'$ are assigned using the same lexical category scheme as Eq.~(2).

%------------------------------------------------------------------------
\subsection{Retrieval and Ranking} 
\label{subsec:retrieval_ranking}

Given the gallery embeddings $\{\mathbf{v}(V)\}_{V\in\mathcal{G}}$ and query embedding $\mathbf{q}(V_r,E)$, we rank all gallery videos by computing their cosine similarities with the query:
\begin{equation}
s(V) = \mathrm{sim}(\mathbf{q}(V_r,E), \mathbf{v}(V)) = \frac{\mathbf{q}(V_r,E)^\top \mathbf{v}(V)}{\|\mathbf{q}(V_r,E)\|\,\|\mathbf{v}(V)\|}.
\end{equation}
Since all embeddings are L2-normalized, the cosine similarity simplifies to the dot product:
\begin{equation}
s(V) = \mathbf{q}(V_r,E)^\top \mathbf{v}(V).
\end{equation}
The gallery videos are ranked in descending order of their similarity scores, and the target video is retrieved as:
\begin{equation}
V_t^* = \operatorname*{arg\,max}_{V\in\mathcal{G}} s(V).
\end{equation}

%% file: sec/4_Experiments.tex
\section{Experiments}
\label{sec:experiments}

\noindent\textbf{Datasets.}We evaluate on two benchmarks that emphasize different aspects of composed video retrieval. \emph{CoVR-R} (ours) contains {2.8k} triplets with structured reasoning annotations and hard distractors, specifically designed to require temporal, causal, and cinematographic reasoning beyond keyword matching. Each triplet includes ground-truth after-effects across five dimensions (states, actions, scene, camera, tempo) as structured reasoning records (Eq.~1). %Triplets are sourced from Dense-WebVid-CoVR~\cite{thawakar2025beyond} and Something-Something V2~\cite{goyal2017something}, filtered through five reasoning-dependency criteria: (i) temporal dependency, (ii) state transition, (iii) cinematography, (iv) implicit cause-effect, and (v) low lexical sufficiency (Sec. ~\ref{sec:covrr_benchmark}). Each assertion undergoes two-stage human verification (Sec.~\ref{sec:human_eval}).%
\emph{Dense-WebVid-CoVR}~\cite{thawakar2025beyond} provides large-scale evaluation with {2.5K} manually curated test triplets from WebVid10M, featuring dense edit texts (avg. 31.2 words) and detailed video descriptions (avg. 81.32 words). %\emph{WebVid-CoVR}~\cite{webvid-covr} offers a standard benchmark with {2.5K} test samples and more concise edits, enabling comparison with prior work. Together, these benchmarks span reasoning-dependent edits, dense modifications, and standard retrieval settings, providing a comprehensive evaluation of our approach.%

\begin{table*}[t]
    \input{tables/main_table}
    \label{tab:main_table}
\end{table*}

\begin{table*}[t]
    \input{tables/dense-webvid-covr}
    \label{tab:dense-webvid-covr}
\end{table*}

\noindent\textbf{Evaluation Metrics.} We evaluate model performance across two complementary dimensions: retrieval accuracy and reasoning quality. 

\noindent\textit{Retrieval Accuracy.} Following standard protocols for composed video retrieval~\cite{cirr,webvid-covr}, we evaluate retrieval performance using Recall@K (R@K), where $K$ denotes the number of top-ranked results considered. Specifically, we report R@1, R@5, R@10, and R@50, which measure the percentage of queries for which the ground-truth target appears within the top-$K$ retrieved results. Higher recall values indicate superior retrieval accuracy.

% can move it supple if needed 
\noindent\textit{Reasoning Score.} To assess whether models correctly predict the causal, temporal, and cinematographic after-effects implied by edits, we introduce a reasoning evaluation framework unique to CoVR-R. Following recent work on structured reasoning assessment~\cite{thawakar2025llamav,dissanayake2025goodfoundationmodelsstepbystep}, we employ an LLM-as-a-judge approach using GPT-4o ($Temp.=0.0$) to compare model-generated reasoning traces against ground truth across ten dimensions. 
% \emph{Faithfulness-Step} (step-level alignment), \emph{Faithfulness-Token} (token-level coherence), \emph{Informativeness-Step} (coverage of key consequences), \emph{Repetition-Token} (absence of redundancy), \emph{Hallucination} (detection of unfounded effects), \emph{Redundancy} (superfluous steps), \emph{Semantic Coverage-Step} (coverage of state/camera/temporal changes), \emph{Reasoning Alignment} (effect chain coherence), \emph{Commonsense} (physical and temporal plausibility), and \emph{Missing Step} (omitted critical effects). 
Each dimension is scored 1-10, and the Overall Reasoning Score is their arithmetic mean. A higher score indicates that the model's reasoning better aligns with the ground truth after-effects reasoning trace. 
% going beyond keyword matching to capture compositional understanding.

\subsection{Implementation Details}
\label{subsec:implementation}

% \noindent\textbf{Video and text representation.}
% We employ a pre-trained vision-language model~\cite{qwen3} (8B parameters) as our multimodal backbone for both gallery encoding and query reasoning. The model remains frozen—no task-specific fine-tuning is performed. For video processing, we uniformly sample frames at 1 fps for both stages. We feed the sampled frames to the model with text prompts to generate contextualized embeddings.

% \noindent\textbf{Embedding function $f$.}
% Our embedding function $f$ extracts token-level embeddings $\{\mathbf{h}_i\}_{i=1}^N$ from the final transformer layer of the vision-language model. These embeddings are aggregated via importance-weighted pooling (Eq.~\ref{eq:weighted_pool}) to produce video representations $\mathbf{v}(V) \in \mathbb{R}^d$.

% \noindent\textbf{Implementation.}
% All experiments are conducted on 8$\times$ NVIDIA A100 GPUs (80GB). We use the HuggingFace Transformers library~\cite{huggingface} for model implementation.

We employ Qwen3-VL~\cite{yang2025qwen3} (8B parameters) as our frozen multimodal backbone without task-specific fine-tuning. Videos are uniformly sampled at 1 fps and processed with task-specific prompts. Our embedding function $\mathbf{v}()$ extracts token-level hidden states $\{\mathbf{h}_i\}_{i=1}^N$ from the penultimate transformer layer and aggregates them via importance-weighted pooling (Eq.~\ref{eq:weighted_pool}) to produce $\mathbf{v}(V) \in \mathbb{R}^{4098}$. Token weights are assigned by semantic informativeness: $\alpha_{\text{high}} = 1.0$ for content-bearing tokens (verbs, nouns, state descriptors), $\alpha_{\text{mid}} = 0.3$ for stop words and generic video terms, and $\alpha_{\text{low}} = 0.1$ for punctuation. All embeddings are L2-normalized. For two-stage reasoning, we generate traces with $T_r=0.8$, top-$p=0.9$ (max 128 tokens), then condition descriptions with $T_d=0.6$ (max 256 tokens). Experiments use single A100 GPUs (40GB) with mixed-precision inference via HuggingFace Transformers~\cite{huggingface_transformers}. Gallery embeddings are precomputed and cached.

\noindent\textbf{Baselines:}
We compare against three primary baselines: CoVR-BLIP~\cite{ventura2024covr}, Thawakar et al.~\cite{thawakar2025beyond}, BSE-CoVR~\cite{thawakar2025beyond}, and video adapted version MVFT-JI$^\dagger$ of \cite{tu2025mllm}. All baselines use BLIP as the backbone with $15$ uniformly sampled frames. We report results for both average pooling (Avg) and cross-attention (CA) fusion strategies.

% \subsection{Results on Composed Video Retrieval (CoVR)}
\subsection{Results}
\noindent\textbf{Results on CoVR-R benchmark:} Table~\ref{tab:covr_r_main} presents results on our proposed \emph{CoVR-R} benchmark, which emphasizes edits where success depends on predicting implicit consequences, a dimension that complements keyword/description-based matching strategies. As large multimodal models demonstrate increasingly strong vision-language alignment, \emph{CoVR-R} provides a critical testbed for evaluating whether these models can reason about temporal causality and cinematographic consequences, capabilities essential for real-world video understanding but not captured by existing retrieval benchmarks.
Prior methods exhibit substantially lower performance on CoVR-R (rows 1-6, avg R@1 = ${32.05}\%$) compared to their reported results on Dense-WebVid-CoVR~\cite{thawakar2025beyond} (rows 4-9, avg R@1 = ${40.66}\%$), indicating that reasoning-dependent edits pose distinct challenges beyond standard retrieval. Our approach without explicit reasoning (row 9) achieves $44.32\%$ R@1, outperforming prior methods by ${+10.89}\%$. Notably, our reasoning-augmented variant (row 10, ${49.88}\%$ R@1) demonstrates the value of explicit after-effect prediction, achieving a reasoning score of {$8.31 \pm 0.098$}. The reasoning traces enable our model to predict state transitions, temporal phases, and cinematographic changes that are only implied by the edit text rather than explicitly stated. The performance gap between \emph{CoVR-R} and standard benchmarks validates that reasoning about causal and temporal consequences provides a complementary signal to surface-level text matching for complex compositional edits.

\noindent\textbf{Results on Dense-WebVid-CoVR test set:} Table~\ref{tab:sota_comparison} presents results on \emph{Dense-WebVid-CoVR}, a large-scale benchmark with detailed edit texts. Our zero-shot approach (row 10) achieves ${58.19\%}$ R@1 and ${97.14\%}$ R@50, substantially outperforming the strongest baseline BSE-CoVR~\cite{thawakar2025beyond} (row 9: $48.08\%$ R@1, $93.78\%$ R@50) by ${+10.11}$ and ${+3.36}$ points, representing relative improvements of ${21.03\%}$ and ${3.58\%}$. Notably, our method excels without video captions or task-specific finetuning. Among caption-free methods, we outperform CoVR-BLIP (row 7: $35.60\%$ R@1) by ${+22.59}$ points (${63.45\%}$ relative improvement), demonstrating that reasoning with a frozen LMM is highly effective without auxiliary supervision. The reasoning-augumented variant (row 11) further improves performance to ${61.21\%}$ R@1 ($+3.02$ percentage points) and $97.61\%$ R@50, achieving a reasoning score of ${8.31 \pm 0.10}$. Both variants substantially exceed all baselines across all metrics, validating the effectiveness of our zero-shot approach on standard CoVR benchmarks. Additional results on \emph{WebVid-CoVR}~\cite{webvid-covr} are provided in the supplementary material.
%The reasoning-augmented variant (row 11) exhibits a modest drop ($-3.02$ R@1, $-2.90$ R@5), suggesting that on this benchmark with more explicit edits, the additional reasoning step may introduce minor noise. However, both variants substantially exceed all baselines across all metrics, validating the effectiveness of our zero-shot approach on standard CoVR benchmarks.

%\begin{table}[t]
%    \input{tables/egocvr}
%    \label{tab:egocvr}
%\end{table}
\begin{table}[t]
    \input{tables/ablations1}
    \label{tab:ablations1}
\end{table}

\begin{table}[t]
    \input{tables/token_aggre}
    \label{tab:token_agre_ablation}
\end{table}

\subsection{Ablation}
To understand our design choices, we conduct ablation studies on our \emph{CoVR-R} benchmark. Please refer suppl. material for additional results and ablations.

\noindent\textbf{Effect of Backbone Scale.} Table~\ref{tab:ablations2} compares different vision-language backbones on CoVR-R while keeping the aggregation strategy fixed (importance-weighted pooling). Performance scales consistently with model capacity: across Qwen3-VL variants, the 72B model achieves $55.48\%$ R@1, the 8B model reaches $49.88\%$, and the 4B model attains $43.98\%$. Comparing across architectures at similar parameter scales, Qwen3-VL models consistently outperform Qwen2.5-VL counterparts (e.g., Qwen3-VL-4B: $43.98\%$ vs. Qwen2.5-VL-3B: $41.78\%$; Qwen3-VL-8B: $49.88\%$ vs. Qwen2.5-VL-7B: $47.56\%$), demonstrating the benefit of improved video understanding and longer context windows in the Qwen3 architecture. 
% Notably, Qwen3-VL-4B outperforms the larger Qwen2.5-VL-3B despite having similar parameter counts, confirming that retrieval improvements stem from architectural advances in temporal reasoning rather than model scale alone. 
These results validate that our zero-shot approach scales effectively across model capacities without task-specific fine-tuning.

\begin{table}[t]
    \input{tables/ablations2}

    \label{tab:ablations2}
\end{table}

% Larger models achieve better retrieval performance through richer video understanding and stronger temporal reasoning capabilities. Qwen3-VL models consistently outperform Qwen2.5-VL counterparts at similar parameter scales, demonstrating the benefit of improved video understanding and longer context windows. Performance scales consistently across Qwen3-VL variants: the 72B model achieves {\color{red}\textbf{55.48}}\% R@1, the 7B model reaches {\color{red}\textbf{49.88}}\%, and even the smallest 3B model attains {\color{red}\textbf{TK}}\%, validating that our zero-shot approach scales effectively across model capacities without task-specific finetuning.

\noindent\textbf{Does Iterative Refinement help?} We investigate whether iteratively refining reasoning traces improves retrieval. In this variant, the model generates an initial reasoning trace describing the after-effects of the edit, then refines it over 5 rounds by re-prompting with the previous trace. The final consolidated trace is used for embedding generation. Table~\ref{tab:ablations1} row 3 shows this yields gains over single-pass reasoning (R@1: 49.88\% $\rightarrow$ 50.56\%) while maintaining stable reasoning quality ($\sim$8.31). However, the 5$\times$ computational overhead makes this impractical for large-scale retrieval. Our final model uses single-pass reasoning for efficiency.

\noindent\textbf{Does Pooling Strategy Matter?} Table~\ref{tab:token_agre_ablation} evaluates pooling strategies for aggregating token embeddings from LMM-generated retrieval embeddings on CoVR-R. A simple {mean pooling} of LMM-generated descriptions is a strong baseline (R@1 {44.87}). {Max} (35.95) loses contextual coherence and {last-token} pooling collapses (1.51), indicating recency bias is inadequate for temporally grounded retrieval; concatenations ({mean$\oplus$last} 14.41, {mean$\oplus$max} 41.77) also underperform. In contrast, a \emph{parameter-free importance-weighted pooling} that down-weights fillers and up-weights action/object/state tokens reaches {49.88} R@1 (+5.01 pp over mean; +11.2\% relative), outperforming all concatenation variants which suggests that targeted semantic emphasis is more effective than adding architectural complexity.

%% file: tables/main_table.tex
% \begin{table*}[t]
    \centering
    % \label{tab:covr_r_main}
    \setlength{\tabcolsep}{8pt}
    \resizebox{\textwidth}{!}{%
    \begin{tabular}{clcccc|cccc|c}
    \toprule
    \rowcolor{gray!15} & & Using Video  & Modification &  &   &  \multicolumn{4}{c}{Recall@K} & Reasoning\\
    \rowcolor{gray!15} & Model &  Captions  & Text Fusion & Backbone &  Frames & R@1 & R@5 & R@10 & R@50 & score \\
    \midrule
    1 & CoVR-BLIP \cite{webvid-covr}  & \ding{56}  & Avg & BLIP & $15$ & $30.30$ & $51.07$ & $57.05$ & $73.82$ & $4.85 \pm 0.12$ \\
    2 & Thawakar \textit{et al.} \cite{thawakar2024composed} & \ding{52}  & Avg & BLIP & $15$ & $31.71$ & $52.18$ & $59.53$ & $74.67$ & $5.62 \pm 0.11$ \\
    3 & BSE-CoVR \cite{thawakar2025beyond}  & \ding{52}   & Avg & BLIP & $15$ & $33.43$ & $53.88$ & $61.31$ & $77.32$ & $6.05 \pm 0.11$ \\
    % \midrule
    4 & CoVR-BLIP \cite{webvid-covr} & \ding{56}  & CA & BLIP & $15$ & $28.06$ & $47.80$ & $55.93$ & $73.76$ & $5.10 \pm 0.13$ \\
    5 & Thawakar \textit{et al.} \cite{thawakar2024composed} & \ding{52}  & CA & BLIP & $15$ & $30.90$ & $50.63$ & $60.11$ & $75.33$ & $5.88 \pm 0.12$ \\
    6 & BSE-CoVR \cite{thawakar2025beyond} & \ding{52}   & CA & BLIP & $15$ & $37.90$ & $57.67$ & $64.48$ & $79.47$ & $6.42 \pm 0.12$ \\
    \dashedmidrule
    7 & MVFT-JI$^{\dagger}$ \cite{tu2025mllm} & \ding{56}   & CA & BLIP & $15$ & $29.48$ & $49.2$2 & $58.02$ & $74.55$ & $6.00 \pm 0.12$ \\
    8 & MVFT-JI$^{\dagger}$ \cite{tu2025mllm} & \ding{52}   & CA & BLIP & $15$ & $34.40$ & $54.15$ & $62.30$ & $77.40$ & $6.28 \pm 0.12$ \\
    \dashedmidrule
    \rowcolor{cyan!5}
    9  & \textbf{Our Approach}   & \ding{56}   & SA & Qwen-VL & FrameR=1 & $\textbf{44.32}$ & $\textbf{61.91}$ & $\textbf{67.33}$ & $\textbf{79.90}$ & $\textbf{7.46 $\pm$ 0.11}$ \\
    \rowcolor{cyan!5}
    10 & \textbf{Our Approach+R} & \ding{56}   & SA & Qwen-VL & FrameR=1 & $\textbf{49.88}$ & $\textbf{66.99}$ & $\textbf{72.97}$ & $\textbf{85.14}$ & $\textbf{8.31 $\pm$ 0.098}$ \\
    \bottomrule
    \end{tabular}
    }
    % \vspace{-1em}
    \caption{
    % \textbf{Comparison of our approach with existing methods on our proposed CoVR-R benchmark}. Our proposed approach consistently outperforms existing methods in \textit{all} settings over Recall@K metrics and reasoning score.
    \textbf{Zero-shot comparison on CoVR-R}. Columns show video caption usage, modification text fusion strategy (Avg vs. CA = cross-attention vs. SA = self-attention),  backbone model, frames per clip, Recall@K, and Reasoning score ($0–10$; mean $\pm$ s.e.m.). MVFT-JI$^\dagger$ denotes our video adaptation of~\cite{tu2025mllm} for fair comparison. Among BLIP-based baselines, stronger fusion and video modeling improve both retrieval and reasoning. Our approach with Qwen3-VL-8B achieves the best zero-shot performance, with substantial gains in R@1--R@50 and reasoning scores without requiring video captions. Our Approach+R further improves results through explicit reasoning.
    }
    \label{tab:covr_r_main}
    \vspace{-0.5em}

%% file: tables/dense-webvid-covr.tex
% \begin{table*}[t]
    \centering
    \setlength{\tabcolsep}{8pt}
    \resizebox{\textwidth}{!}{%
    \begin{tabular}{clclccc|cccc}
        \toprule
        \rowcolor{gray!15} & & Using Video  &  & Modification &  &   &  \multicolumn{4}{c}{Recall@K}  \\
        \rowcolor{gray!15} & Model &  Captions & Input Modalities & Text Fusion & Backbone &  Frames & R@1 & R@5 & R@10 & R@50 \\
        \midrule
        1 & Random & & - & - & - & - & $0.04$ & $0.21$ & $0.32$ & $1.46$ \\
        2 & CoVR-BLIP \cite{webvid-covr}  & \ding{56} & Text & - & BLIP & - & $24.12$ & $56.02$ & $60.16$ & $82.34$ \\
        3 & CoVR-BLIP \cite{webvid-covr}  & \ding{56} & Visual & - & BLIP & $15$ & $22.52$ & $53.08$ & $58.34$ & $81.26$ \\
        % 4 & CoVR-BLIP  & \ding{56} & Visual + Text & Avg & CLIP & 15 & 44.37 & 69.13 & 77.62 & 93.00 \\
        4 & CoVR-BLIP \cite{webvid-covr}  & \ding{56} & Visual + Text & Avg & BLIP & $15$ & $38.44$ & $64.96$ & $71.72$ & $87.12$ \\
        5 & Thawakar .\textit{et.al.} \cite{thawakar2024composed} & \ding{52} & Visual + Text & Avg & BLIP & $15$ & $40.23$ & $66.38$ & $74.84$ & $88.12$ \\
        6 & BSE-CoVR \cite{thawakar2025beyond}  & \ding{52}  & Visual + Text & Avg & BLIP & $15$ & $42.41$ & $68.54$ & $77.07$ & $91.24$ \\
        % \midrule
        7 & CoVR-BLIP \cite{webvid-covr} & \ding{56} & Visual + Text & CA & BLIP & $15$ & $35.60$ & $60.80$ & $70.31$ & $87.05$ \\
        8 & Thawakar .\textit{et.al.} \cite{thawakar2024composed} & \ding{52} & Visual + Text & CA & BLIP & $15$ & $39.20$ & $64.40$ & $75.56$ & $88.90$ \\
        9 & BSE-CoVR \cite{thawakar2025beyond} & \ding{52}  & Visual + Text & CA & BLIP & $15$ & $48.08$ & $73.36$ & $81.06$ & $93.78$ \\
        \dashedmidrule
        \rowcolor{cyan!5}
        10 & \textbf{Our Approach} & \ding{56}  & Visual + Text & SA & Qwen-VL & FrameR=1 & \textbf{58.19} & \textbf{80.50} & \textbf{86.92} & \textbf{97.14} \\
        \rowcolor{cyan!5}
        11 & \textbf{Our Approach+R} & \ding{56}  & Visual + Text & SA & Qwen-VL & FrameR=1 & \textbf{61.21} & \textbf{83.40} & \textbf{89.39} & \textbf{97.61} \\
        \bottomrule
    \end{tabular}
    }
    % \vspace{-0.8em}
    % \caption{\textbf{Comparison of our approach with existing methods on the Dense-WebVid-CoVR test set}. Our proposed approach consistently outperforms existing methods in \textit{all} settings and Recall@K metrics. %By leveraging detailed video descriptions and dense modification texts, our method consistently achieves higher recall scores across all Recall@K metrics. 
    % Notably in the Visual + Text setting with Cross-Attention (CA), our method improves Recall@1 to 71.26 and Recall@50 to 98.88. Best results are in bold.}
    \caption{\textbf{Results on \emph{Dense-WebVid-CoVR} test set.} Our approach with Qwen3-VL-8B (rows $10-11$) outperforms all prior baselines. Row 10 achieves $61.21$\% R@1 and $97.61$\% R@50, surpassing the strongest baseline BSE-CoVR (row 9: $48.08$\% R@1, $93.78$\% R@50) by $+13.13$ and $+3.83$ points. Row 11 (+R) incorporates explicit reasoning traces. \ding{52}: uses video captions; \ding{56}: caption-free. CA: Cross-Attention;SA: Self Attention; Avg: Average pooling. Best results in \textbf{bold}.}
    \label{tab:sota_comparison}
    % \vspace{-1em}
% \end{table*}

%% file: tables/ablations1.tex
% \begin{table*}[t]
    \centering
    \setlength{\tabcolsep}{4pt}
    \resizebox{\linewidth}{!}{%
        \begin{tabular}{l|cccc|c}
        \toprule
            Method & R@1 & R@5 & R@10 & R@50 & Reasoning Score \\
            \toprule
             \rowcolor{cyan!5}
             Qwen3-VL-8b & $44.32$ & $61.91$ & $67.33$ & $79.90$ & -- \\ 
             \rowcolor{cyan!5}
              $+$ Reasoning & $49.88$ & $66.99$ & $72.97$ & $85.14$ & $8.31$ ± $0.0980$ \\  
             \rowcolor{cyan!5}
             $+$  Iterative Refinement & $50.56$ & $74.03$ & $81.24$ & $92.17$ & $8.31$ ± $0.1011$ \\   
             % Qwen3-VL-8b + GT Reasoning & -- & -- & -- & -- & -- \\   
            \bottomrule
        \end{tabular}
        }
        % \vspace{-1em}
        \caption{
        \textbf{Effect of refined reasoning on retrieval}.%  We report Recall@K and a Reasoning Score (0--10; checklist satisfaction, mean $\pm$ s.e.m.). 
        Starting from Qwen3-VL-8B, adding our Reasoning module lifts retrieval (R@1 $44.32\%$ $\rightarrow$ $49.88\%$) and improves the Reasoning Score. Iterative Refinement further boosts performance (R@1=$50.56\%$) while maintaining reasoning quality ($\sim$8.31), but incurs 5$\times$ inference cost. Our final model uses single-pass reasoning.}
% \end{table*}

%% file: tables/token_aggre.tex
\centering
\setlength{\tabcolsep}{8pt}
\resizebox{\linewidth}{!}{%
    \begin{tabular}{l|ccccc}
    \toprule
    \rowcolor{gray!5}
    \textbf{Strategy} & \textbf{R@1} & \textbf{R@5} & \textbf{R@10} & \textbf{R@50} & \textbf{R$_{\text{mean}}$} \\
    \midrule
     Last & $1.51$ & $3.57$ & $4.64$ & $10.14$ & $3.24$ \\ 
     Mean$\oplus$Last & $14.41$ & $25.11$ & $29.20$ & $41.77$ & $22.91$ \\
     Max & $35.95$ & $52.02$ & $72.05$ & $93.98$ & $48.78$ \\ 
     Mean$\oplus$Max & $41.77$ & $58.84$ & $65.41$ & $77.32$ & $55.34$ \\ 
     Mean & $44.87$ & $63.67$ & $69.61$ & $82.44$ & $59.39$ \\ 
     %Uniform & 41.8494 & 60.3436 & 67.1418 & 81.3596 & 56.4449 \\ 
     %spaCyStopWords & 45.8333 & 64.0716 & 70.2120 & 83.8816 & 60.0390 \\ 
     %NLTKStopWords & 45.8333 & 64.0716 & 70.2120 & 83.8816 & 60.0390 \\ 
     %Graduated & 46.8933 & 65.0950 & 71.6374 & 83.9912 & 61.2086 \\ 
    % \midrule
    \rowcolor{cyan!5}
     Weighted & $49.88$ & $66.99$ & $72.9$ & $85.14$ & $63.28$ \\ 
    \bottomrule
    \end{tabular}
}
% \vspace{-1em}
\caption{\textbf{Token aggregation strategies (\emph{CoVR-R}).} We evaluate pooling methods for aggregating token embeddings in our reasoning model. Importance weighting achieves the best performance.}
\label{tab:ablation_pooling_v2}
\vspace{-1em}

%% file: tables/ablations2.tex
% \begin{table*}[t]
    \centering
    \setlength{\tabcolsep}{5pt}
    \resizebox{\linewidth}{!}{%
        \begin{tabular}{l|cccc|c}
        \toprule
            Method & R@1 & R@5 & R@10 & R@50 & Reasoning Score \\
            \toprule
            \rowcolor{cyan!5}
            Qwen3-VL-4B      & $43.98$ & $61.89$ & $68.27$ & $80.69$ & $7.95 \pm 0.11$ \\
            \rowcolor{cyan!5}
            Qwen3-VL-8B      & $\mathbf{49.88}$ & $\mathbf{66.99}$ & $\mathbf{72.97}$ & $\mathbf{85.14}$ & $\mathbf{8.31 \pm 0.098}$ \\
            % \rowcolor{cyan!5}
            % Qwen3-VL-32b   & -- & -- & -- & -- & -- \\
            \rowcolor{cyan!5}
            Qwen3-VL-72B     & $55.48$ & $72.69$ & $78.57$ & $87.99$ & $9.05 \pm 0.09$ \\
            \rowcolor{cyan!5}
            \midrule
            Qwen2.5-VL-7B    & $47.56$ & $64.08$ & $69.84$ & $82.33$ & $8.06 \pm 0.10$ \\
            \rowcolor{cyan!5}
            Qwen2.5-VL-3B    & $41.78$ & $59.59$ & $66.17$ & $78.59$ & $7.72 \pm 0.11$ \\
            \rowcolor{cyan!5}
            \bottomrule
        \end{tabular}
        }
        % \vspace{-1em}
        \caption{
        \textbf{Impact of model scaling}. Recall@K and Reasoning Score (0–10; checklist satisfaction, mean ± s.e.m.) across Qwen variants. Performance improves with scale: Qwen3-VL-72B is best (R@1=$55.48\%$, R@50=$87.99\%$, Reasoning $9.05 \pm 0.09$), while Qwen3-VL-8B is the strongest mid-size model (R@1=$49.88\%$). Overall, Qwen3 series outperforms comparable Qwen2.5 models on both retrieval and reasoning quality.
    }
% \vspace{-1em}
% \end{table*}

% % \begin{table*}[t]
% \centering
% \setlength{\tabcolsep}{12pt}
% \resizebox{\linewidth}{!}{%
% \begin{tabular}{l|cccc|c}
% \toprule
% Method & R@1 & R@5 & R@10 & R@50 & Reasoning Score \\
% \toprule
% \rowcolor{cyan!5}
% Qwen3-VL-3b      & $43.98$ & $61.89$ & $68.27$ & $80.69$ & $7.95 \pm 0.11$ \\
% \rowcolor{cyan!5}
% Qwen2.5-VL-7b    & $47.56$ & $64.08$ & $69.84$ & $82.33$ & $8.06 \pm 0.10$ \\
% \rowcolor{cyan!5}
% Qwen2.5-VL-3b    & $41.78$ & $59.59$ & $66.17$ & $78.59$ & $7.72 \pm 0.11$ \\
% \rowcolor{cyan!5}
% Qwen3-VL-8b      & $\mathbf{49.88}$ & $\mathbf{66.99}$ & $\mathbf{72.97}$ & $\mathbf{85.14}$ & $\mathbf{8.31 \pm 0.098}$ \\
% % \rowcolor{cyan!5}
% % Qwen3-VL-32b   & -- & -- & -- & -- & -- \\
% \rowcolor{cyan!5}
% Qwen3-VL-72b     & $55.48$ & $72.69$ & $78.57$ & $87.99$ & $9.05 \pm 0.09$ \\
% \bottomrule
% \end{tabular}
% }
% \caption{\textbf{Effect of changing the Qwen-VL backbone (CoVR-R).} Qwen3-VL-8B is the measured base; other entries are capacity- and generation-aware expectations under the same pipeline (Qwen3$\!>\!$Qwen2.5; larger$\!>\!$smaller).}
% % \end{table*}

%% file: sec/5_Conclusion.tex
\section{Conclusion}
We presented a \emph{reason–then–retrieve} framework for composed video retrieval that explicitly predicts and verifies the \emph{after–effects} of an edit covering state transitions, action phases, scene context, camera/shot, and tempo. This simple, zero-shot design consistently improves Recall@K and reasoning quality over strong baselines. To fairly evaluate this capability, we introduced \textsc{CoVR-R}, a 2.8k-triplet benchmark with grounded, schema-constrained reasoning traces and hard distractors that mimic real-world queries where keyword overlap is insufficient. Together, our method and benchmark encourage the community to treat reasoning as a first-class objective in CoVR, reduce reliance on task-specific supervision by leveraging general-purpose LMMs, and move retrieval systems toward precise, explainable video search.

%% file: sec/X_suppl.tex
\clearpage
\setcounter{page}{1}
\maketitlesupplementary

\setcounter{figure}{0}
\setcounter{table}{0}
\setcounter{section}{0}
\renewcommand{\theequation}{\Alph{equation}}
\renewcommand{\thefigure}{\Alph{figure}}
\renewcommand{\thesection}{\Alph{section}}
\renewcommand{\thetable}{\Alph{table}}

% \begin{figure*}[ht!]
% \includegraphics[width=\textwidth]{assets/intro_figure.pdf}
% \vspace{-2em}
% \caption{
% \textbf{\textcolor{red}{TODO Why reasoning is needed for CoVR}}. Each row pairs a reference video and edit text with the desired target (highlighted in green), where success depends on after-effects rather than keyword overlap. Examples include: shifting from a brown cow to a solitary brown horse in rolling hills with trees (state and scene changes); switching to a close-up of four hands on a piano (cinematography and object configuration); moving from typing to clenched fists and closing the laptop (temporal phase progression). These cases require inferring consequences such as state transitions, phase order, and shot scale that not merely matching words. Reasoning bridges what the edit says and what the target must show, capturing the visual consequences that define correct retrieval. More examples are in suppl. material.
% }
% \label{fig:supple_figure}
% \vspace{-1em}
% \end{figure*} 

\section{Supplementary Materials}

In this supplement, we provide: (1) additional experimental results on WebVid-CoVR and reasoning granularity ablations (Sec.~\ref{sec:additional_results}); (2) comprehensive dataset statistics and annotation analysis (Sec.~\ref{sec:dataset_stats}); and (3) qualitative analysis including success cases, failure cases, and reasoning trace examples (Sec.~\ref{sec:qualitative}). We hope this document complements our main paper.

For sections, figures, and tables, we use numbers (\eg, Sec.\ 1) to refer to the main paper and capital letters (\eg, Sec.\ A) to refer to this supplement.

\section{Additional Results}
\label{sec:additional_results}

\noindent\textbf{Results on WebVid-CoVR test set.}
Table~\ref{tab:webvid} presents results on the WebVid-CoVR benchmark. Our approach achieves 49.15\% R@1, outperforming the strongest baseline BSE-CoVR~\cite{webvid-covr} (48.08\%) by +1.07 pp, as well as CoVR-BLIP-2~\cite{webvid-covr} (45.66\%) and CoVR-BLIP (45.46\%). However, this R@1 advantage does not extend to higher recall metrics: our R@5 (70.72\%) and R@10 (79.25\%) trail BSE-CoVR~\cite{thawakar2025beyond} (73.36\% R@5, 81.06\% R@10) and CoVR-BLIP-2 (71.71\% R@5, 81.30\% R@10).

\begin{figure}[t]
\centering
\includegraphics[width=\linewidth]{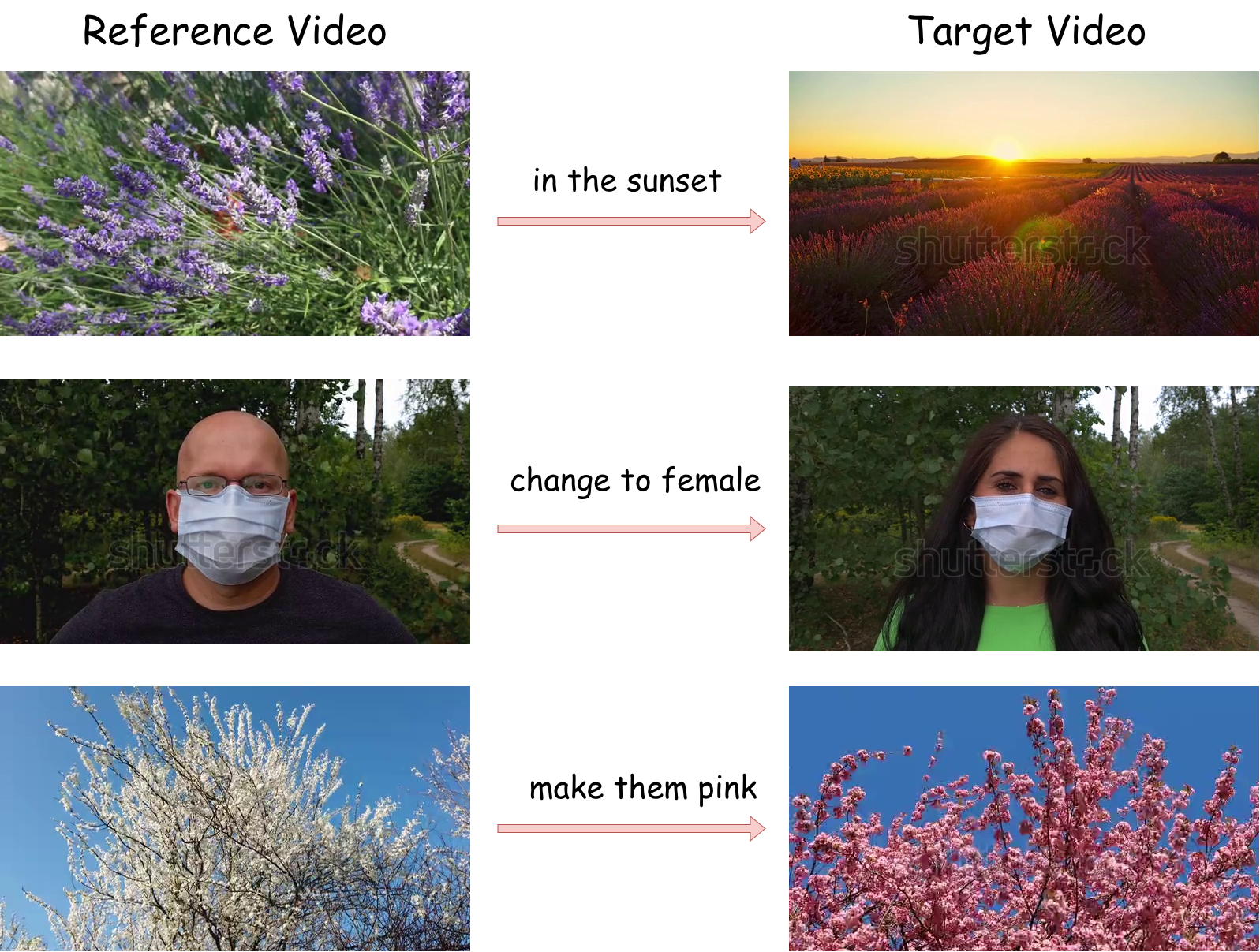}
% \vspace{-2em}
\caption{The figure shows the middle frame from each reference and target video pair in the WebVid-CoVR dataset. The arrows indicate the intended modifications described by the text instructions. The targets reveal that the edit instructions are highly generic, making the composed video retrieval task ambiguous and challenging. } % The dataset favours key word based matching over reasoning.}
\label{fig:webvid_covr_test_bad_sample}
\vspace{-1em}
\end{figure}

We attribute this trade-off to dataset characteristics. As illustrated in Figure~\ref{fig:webvid_covr_test_bad_sample}, WebVid-CoVR predominantly features underspecified modification texts (\eg, ``change to female'', ``make them pink'') where keyword-based matching is highly effective. Many edits lack explicit visual targets (\eg, ``in the sunset'' applied to a close-up flower shot, where the target video shows a distant view making the intended transformation ambiguous). Our reasoning-first approach, designed for complex edits requiring causal inference about implicit after-effects, introduces unnecessary computational overhead when simple keyword matching suffices. The multi-stage process generating after-effect traces and synthesizing target descriptions dilutes discriminative signals compared to direct edit-video alignment for straightforward transformations. This suggests that \emph{reasoning-aware retrieval provides greatest value when edit complexity exceeds surface-level matching}, highlighting the need for adaptive strategies that dynamically select between reasoning and non-reasoning models based on query characteristics.

%Table ~\ref{tab:webvid} presents results on WebVid-CoVR, which features more concise edit texts compared to Dense-WebVid-CoVR. Our approach achieves 49.15\% R@1, underperforming the strongest baseline, \cite{thawakar2024composed} (60.12\% R@1). This preformance gap reflects a fundamental trade-off in our design: while our reasoning-first approach excels at predicting implicit after-effects for complex, underspecified edits (as demonstrated on CoVR-R and Dense-WebVid-CoVR), it incurs overhead when edits are simple and explicit. WebVid-CoVR's concise modifications (e.g., "change background to beach") often directly specify the visual target without requiring causal inference, making keyword-based matching highly effective. In such cases, our multi-stage reasoning process, generating after-effect traces and synthesizing target descriptions, may intriduce noise or dilute discriminative signal compared to direct edit-video alignment. This suggests that reasoning-aware retrieval provides the greatest value when edit complexity exceeds what surface-level matching can resolve, highlighing a promising direction for adaptive retrieval strategies that modulate reasoning depth based on edit characteristics.

\begin{table}[t]
    \input{tables/webvid}
    \label{tab:webvid}
\end{table}

% \begin{table}[t]
%     \input{tables/ablations2}
%     \label{tab:ablations2}
% \end{table}

%\noindent\textbf{Results on CIRR:}

\begin{table*}[t]
    \centering
    \setlength{\tabcolsep}{15pt}
    \resizebox{\linewidth}{!}{%
        \begin{tabular}{l|c|cccc|c|c}
        \toprule
             \textbf{Strategy} & \textbf{Avg. Trace Length} & \textbf{R@1} & \textbf{R@5} & \textbf{R@10} & \textbf{R@50} & \textbf{Reasoning Score} & \textbf{Inference Time} \\
        \midrule
             Minimal (keywords only) & 15 tokens & 44.10 & 59.30 & 65.82 & 77.45 & 6.80 $\pm$ 0.13 & 1.0$\times$ \\
             Compact (atomic assertions) & 45 tokens & 47.80 & 64.70 & 70.15 & 82.33 & 7.90 $\pm$ 0.11 & 1.3$\times$ \\
            \rowcolor{cyan!5}
             \textbf{Standard (ours)} & \textbf{89 tokens} & \textbf{49.88} & \textbf{66.99} & \textbf{72.97} & \textbf{85.14} & \textbf{8.31 $\pm$ 0.098} & \textbf{1.8$\times$} \\
             Verbose (detailed chains) & 186 tokens & 48.20 & 65.10 & 71.34 & 83.67 & 8.42 $\pm$ 0.095 & 3.2$\times$ \\
        \bottomrule
        \end{tabular}
    }
    % \vspace{-1em}
    \caption{\textbf{Impact of reasoning granularity on retrieval performance and efficiency}. We evaluate four levels of reasoning detail on CoVR-R. Standard granularity (89 tokens) achieves the best balance between retrieval accuracy (R@K) and reasoning quality, while verbose traces (186 tokens) improve reasoning scores but degrade retrieval by introducing excessive detail that dilutes discriminative signals.}
    \label{tab:reasoning_granularity}
\end{table*}

\noindent\textbf{Reasoning granularity ablation.}  Table~\ref{tab:reasoning_granularity} investigates how the level of detail in reasoning traces affects both retrieval performance and computational efficiency. We evaluate four granularity levels on CoVR-R: minimal keyword-based traces (avg 15 tokens), compact atomic assertions (avg 45 tokens), our standard structured reasoning (avg 89 tokens), and verbose detailed chains (avg 186 tokens). 

The results reveal a critical trade-off between reasoning detail and retrieval effectiveness. While minimal traces (avg 15 tokens) provide computational efficiency (1.0$\times$ baseline), they achieve only 44.10\% R@1, indicating that oversimplified reasoning fails to capture the causal and temporal consequences necessary for accurate retrieval. Compact traces (45 tokens) improve R@1 to 47.80\%, demonstrating the value of atomic effect predictions, yet still underperform compared to our standard approach.

Our standard granularity (89 tokens) achieves the optimal balance, reaching 49.88\% R@1 and 8.31$\pm$0.098 reasoning score with 1.8$\times$ computational cost. Notably, further increasing detail to verbose traces (186 tokens) yields a counterintuitive result: reasoning quality marginally improves (8.42$\pm$0.095), yet retrieval accuracy \textit{degrades} across all metrics (R@1: 48.20\%, R@5: 65.10\%). This performance degradation, despite nearly doubling the inference cost to 3.2$\times$, suggests that \emph{excessive granularity introduces noise that dilutes discriminative signals} rather than enhancing semantic understanding.

This finding has important implications: the optimal reasoning representation for retrieval is not simply ``more detailed'' but rather one that captures essential after-effects state transitions, temporal phases, and cinematographic cues without overwhelming the embedding space with auxiliary information. Our standard structured reasoning trace achieves this balance by maintaining structured slot constraints (states, actions, scene, camera, tempo) that enforce semantic organization while preventing verbose elaboration. This supports our hypothesis that \emph{reasoning-aware retrieval benefits from constrained, schema-guided effect prediction} rather than unconstrained natural language generation.

\section{Dataset Statistics}
\label{sec:dataset_stats}

We present comprehensive statistics of \emph{CoVR-R}, highlighting its unique multi-level annotation structure. Our dataset provides three complementary annotation types: modification instructions, detailed causal reasoning, and brief summaries which enables both retrieval evaluation and reasoning quality assessment.

\begin{figure}[t]
\centering
\includegraphics[width=0.85\linewidth]{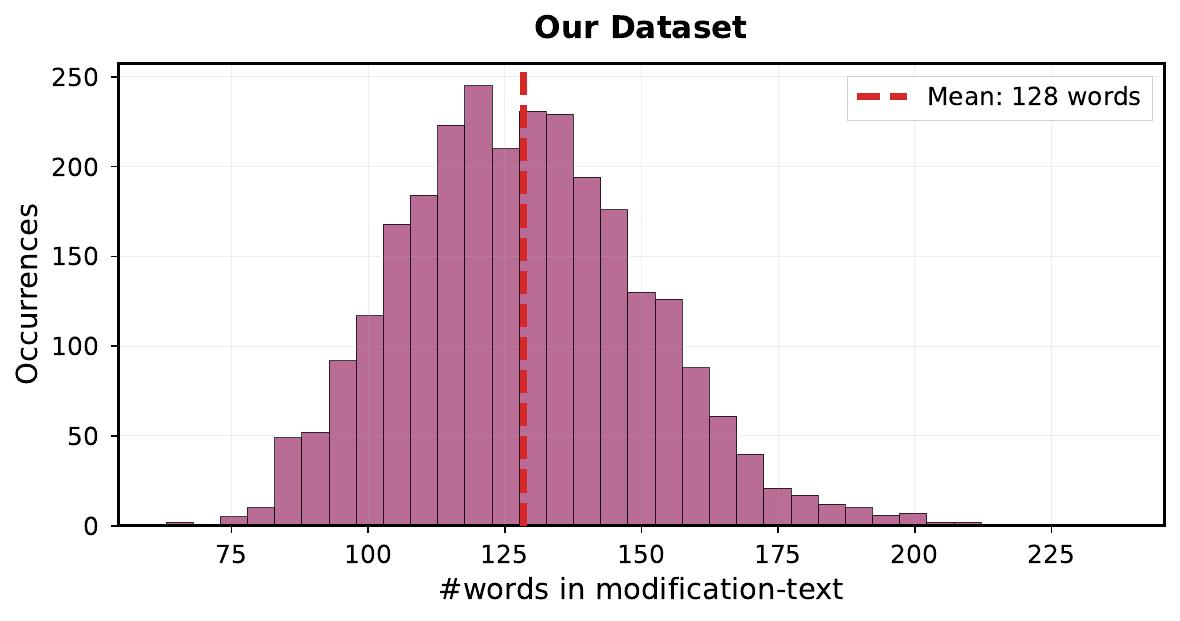}
% \vspace{-2em}
\caption{Distribution of \texttt{modification\_text} in CoVR-R. Mean: 128$\pm$23 words (range: 63-237). These detailed instructions specify not just what changes, but how transformations manifest visually.}
\label{fig:modification_dist}
\vspace{-1em}
\end{figure}

\begin{figure}[t]
\centering
\includegraphics[width=0.85\linewidth]{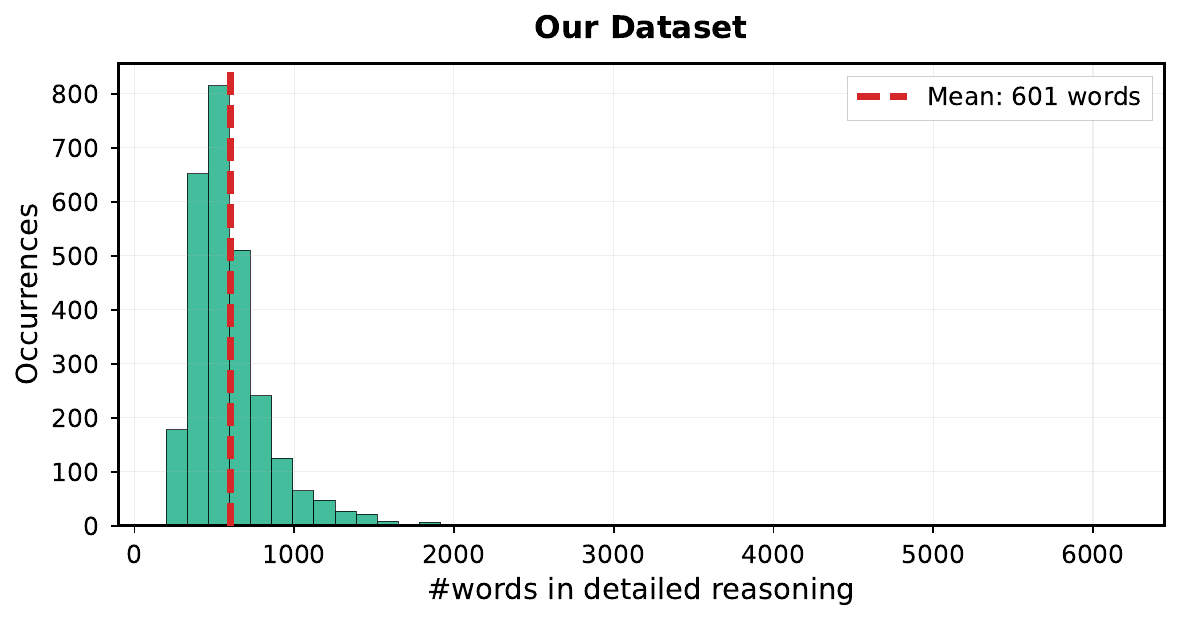}
% \vspace{-2em}
\caption{Distribution of \texttt{reasoning\_detailed} in CoVR-R. Mean: 601 $\pm$ 281 words (range: 197-6150). These comprehensive causal explanations are unique to our work and provide temporal dependencies, state transitions, and compositional relationships.}
\label{fig:detailed_reasoning}
\vspace{-1em}
\end{figure}

\begin{figure}%[t]
\centering
\includegraphics[width=0.85\linewidth]{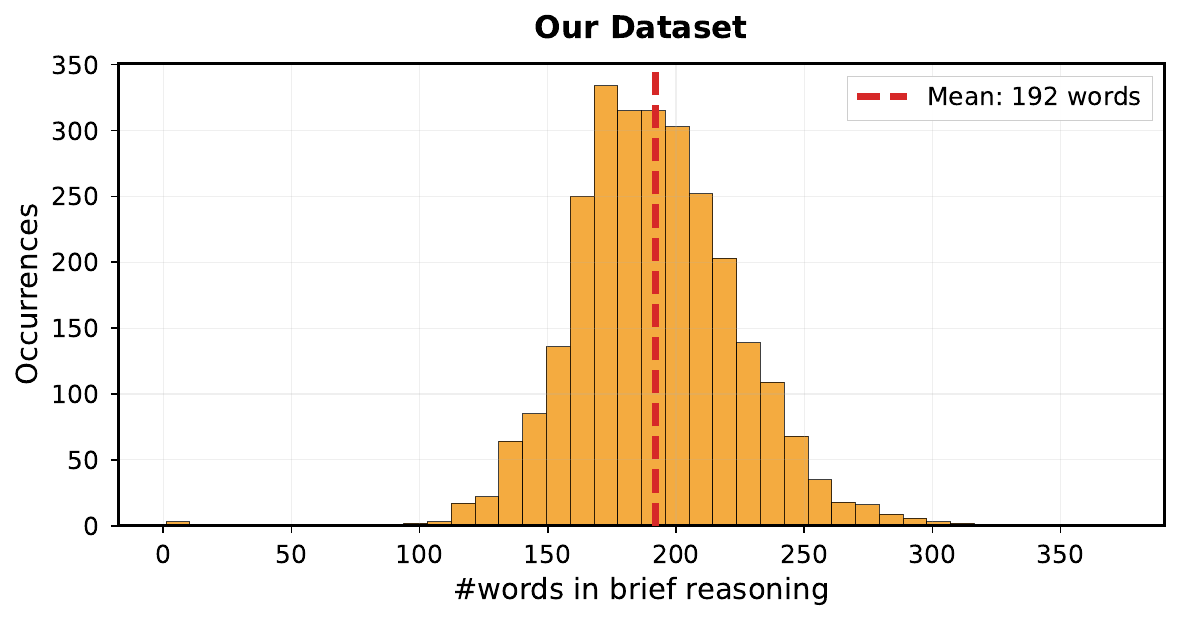}
% \vspace{1em}
\caption{Distribution of \texttt{reasoning} (brief summaries) in our dataset. Mean: 192 $\pm$ 32 words (range: 1-372). These concise overviews bridge modification instructions and detailed reasoning.}
\label{fig:brief_reasoning}
\vspace{-1em}
\end{figure}

\subsection{Annotation Length Analysis}

To quantify the richness of our annotations, we compare the distribution of the number of words in modification texts across three datasets: WebVid-CoVR~\cite{ventura2024covr}, Dense-WebVid-CoVR~\cite{thawakar2025beyond}, and our proposed dataset CoVR-R. In the WebVid-CoVR dataset, the majority of modification texts are very brief, with a word count predominantly ranging between 3 and 7 words, as illustrated in Figure~\ref{fig:webvid_covr_test_bad_sample}. This brevity can limit the models ability to capture the distinct modifications needed for fine-grained video retrieval. Dense-WebVid-CoVR improved upon this with modification texts centered around 30 to 50 words, providing more detailed transformation specifications. In contrast, our dataset provides significantly more detailed modification texts (\texttt{modification\_text}), with the distribution centered around 128 words per instruction (Figure~\ref{fig:modification_dist}). This is approximately 2.5$\times$ to 4$\times$ longer than Dense-WebVid-CoVR and over 18$\times$ longer than WebVid-CoVR. % This increased detail captures how transformations manifest visually, including contextual cues essential for reasoning-aware retrieval.

Beyond modification texts, our dataset uniquely provides two levels of reasoning annotations absent in prior work. \textbf{Detailed Reasoning} (\texttt{reasoning\_detailed}, Figure~\ref{fig:detailed_reasoning}) averages 601 words (range: 197--6,150), explaining the causal relationships, temporal dynamics, and compositional changes underlying each transformation. These annotations enable evaluation of not just \textit{what} changes, but \textit{why} and \textit{how} through explicit reasoning about state transitions and temporal phases. \textbf{Brief Reasoning} (\texttt{reasoning}, Figure~\ref{fig:brief_reasoning}) averages 192 words, providing concise summaries of key transformation effects.

This three-tier structure: modification instructions (128 words), detailed reasoning (601 words), and brief summaries (192 words) totals 922 words per sample, substantially exceeding prior work. While Dense-WebVid-CoVR provides richer descriptions (80--100 words) and longer modifications (30--50 words), our dataset adds explicit causal reasoning chains. This hierarchical design supports multi-level evaluation: retrieval accuracy (modification-based) and reasoning quality (detailed trace alignment). CoVR-R thus serves as the first benchmark for evaluating fine-grained temporal reasoning capabilities in composed video retrieval.

 % and effect prediction (brief summary consistency).
 
% Therefore, CoVR-R serves as a strong benchmark because it is the first to combine substantial annotation depth (922 words/sample), explicit causal reasoning chains absent in prior work, multi-level evaluation of both retrieval accuracy and reasoning quality, and cross-domain complexity—collectively enabling rigorous assessment of fine-grained temporal reasoning capabilities in composed video retrieval.

\subsection{Modification Lexicon}
\begin{figure}[t]
    \centering
    \includegraphics[width=0.95\linewidth]{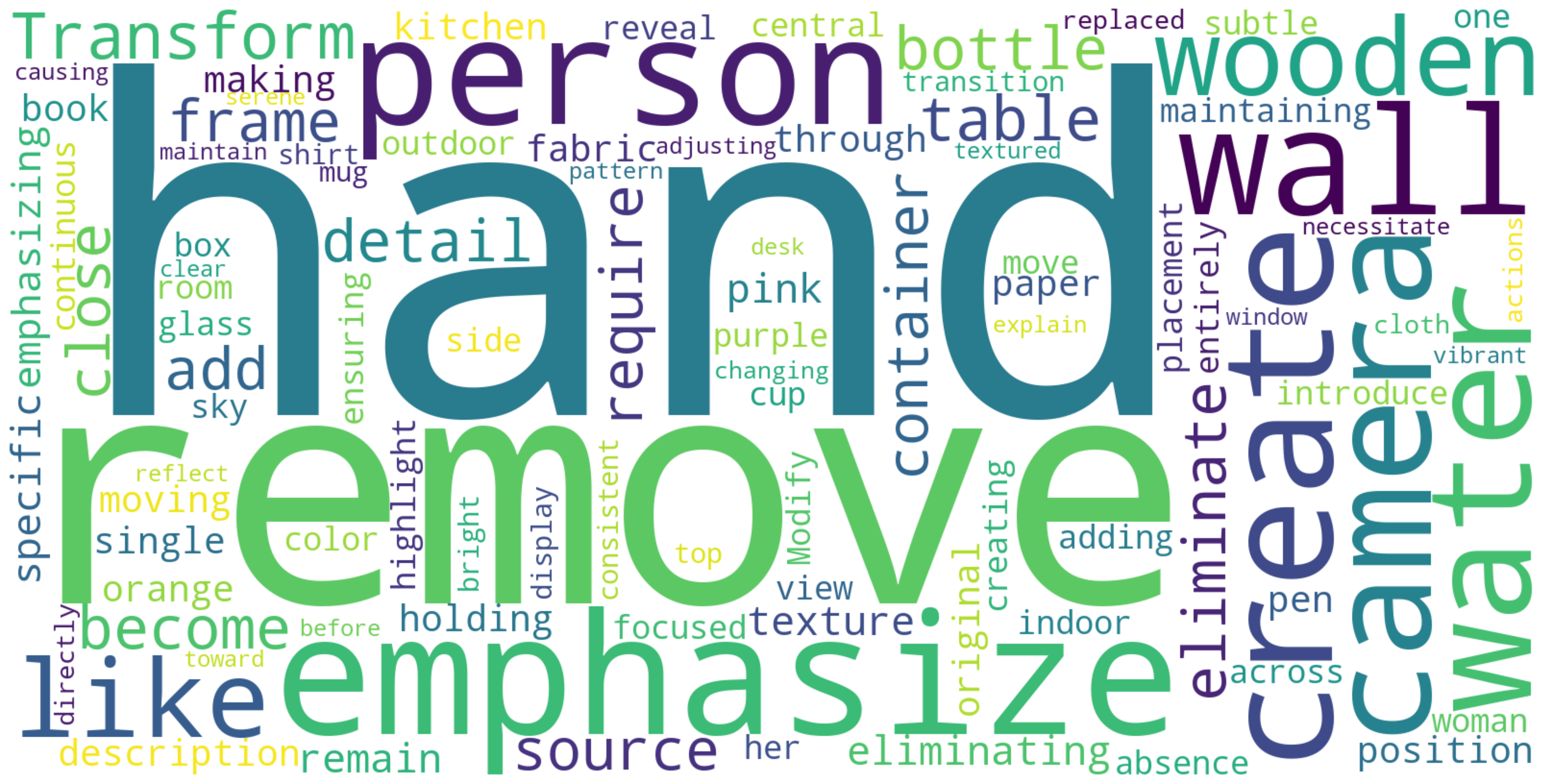}
    \caption{Word cloud visualization of the most frequent content terms in CoVR-R modification texts. Larger words indicate higher frequency and reflect the dataset's diverse, multi-domain nature, with terms like \textit{hand}, \textit{water}, \textit{wooden}, \textit{person}, and \textit{kitchen} highlighting rich contextual variety spanning everyday activities, object interactions, and environmental settings across domains.}
    \label{fig:wordcloud}
\end{figure}

Figure~\ref{fig:wordcloud} presents a word cloud visualization of the most frequently occurring content terms in the modification texts of CoVR-Reasoning. The lexicon reveals the dataset's cross-domain coverage, with prominent terms spanning object interactions (\textit{hand}, \textit{holding}, \textit{bottle}), environmental contexts (\textit{kitchen}, \textit{outdoor}, \textit{water}), materials (\textit{wooden}, \textit{fabric}), human elements (\textit{person}, \textit{camera}), and more.

This diverse vocabulary reflects everyday scenarios and requires models to understand causal relationships between objects, actions, and contexts. The modification texts describe \textit{what changes} (averaging 128 words), while the detailed reasoning annotations (601 words avg) explain \textit{why and how} these changes occur through temporal causality and compositional relationships. Our dataset enables comprehensive evaluation of reasoning capabilities beyond surface-level retrieval.

\section{Qualitative Analysis}
\label{sec:qualitative}

We present qualitative examples illustrating when and why our reasoning-first approach succeeds or fails. These cases reveal critical insights about the relationship between edit complexity, reasoning granularity, and retrieval performance, highlighting both the strengths of explicit causal reasoning and the challenges posed by hyper-specific instructions.

\begin{figure*}[t]
    \centering

    \includegraphics[width=\textwidth]{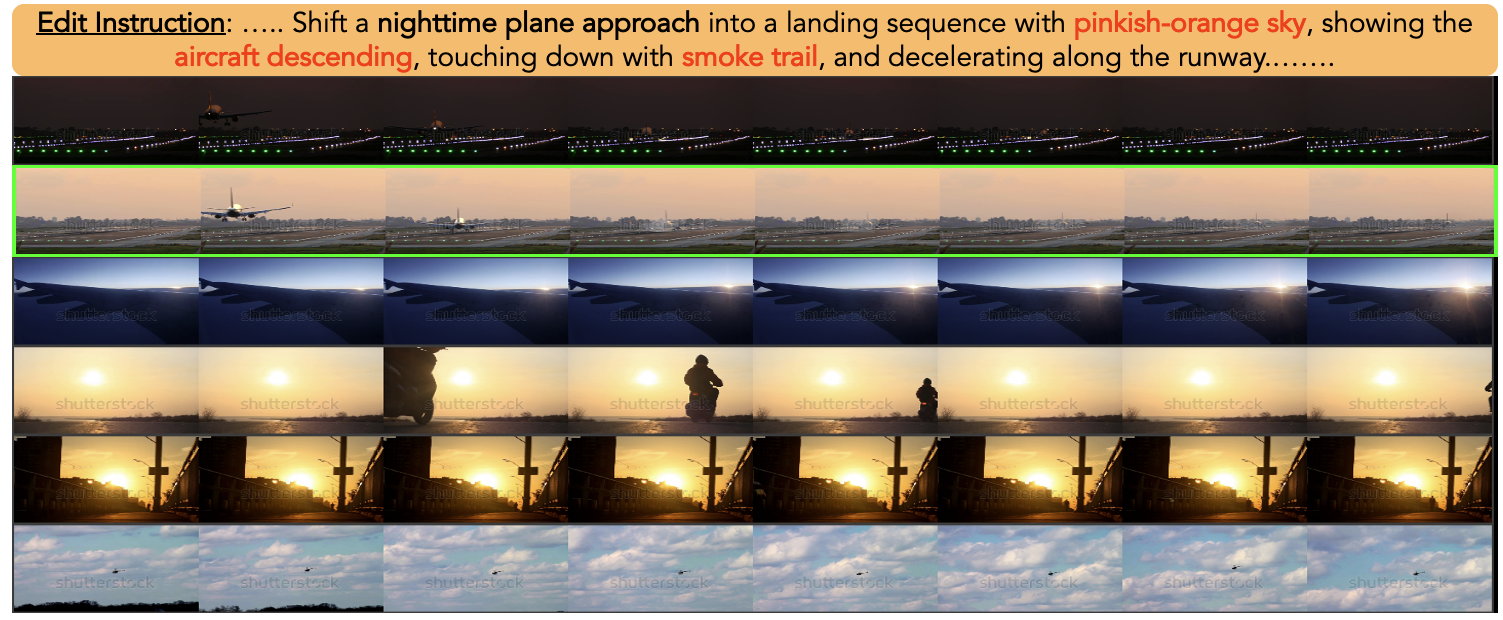}
    \includegraphics[width=\textwidth]{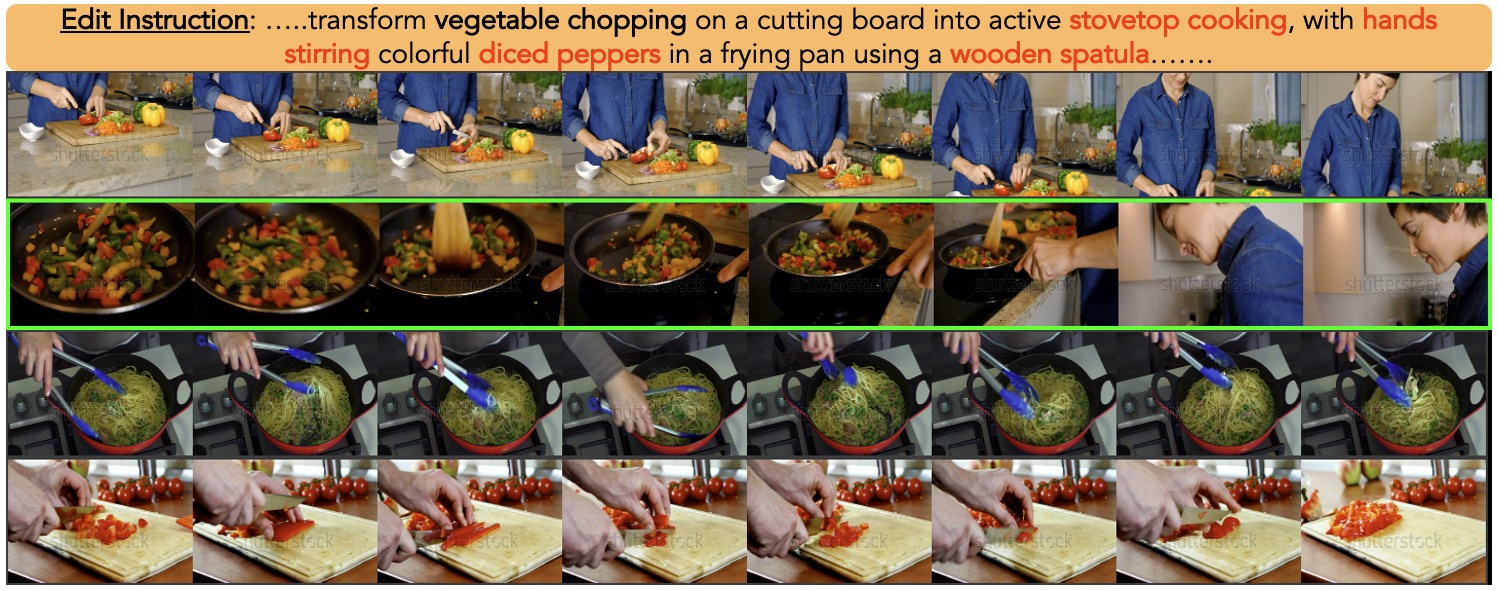}
    \caption{The figure shows two successful retrieval examples where ground truth ranks first:  airplane landing (top) and vegetable cooking  (bottom) . Each example displays: detailed edit instruction in a text box with key concepts highlighted in red, (Row 1) input reference video, (subsequent rows) top-k retrieved predictions with green border indicating the correctly retrieved ground truth at Top-1. Notably, all top retrievals exhibit highly similar visual contexts, demonstrating the effectiveness of LLM-based reasoning in capturing semantic after-effects for composed video retrieval.}

    \label{fig:success}
\end{figure*}

\subsection{Success Case Analysis}

% our Reason-then-Retrieve framework successfully identifies the correct target through the following two-stage process.

Figure~\ref{fig:success} demonstrates how our architecture successfully handles complex implicit transformations. Consider the vegetable cooking example: the edit requires transitioning from "chopping vegetables on a wooden cutting board" to "stirring diced red, green, and yellow peppers in a frying pan on a stovetop, using a close-up shot." This transformation demands reasoning about implicit visual after-effects \emph{what} happens between chopping and stirring, \emph{how} objects change state, and \emph{how} the scene context evolves. A keyword-based baseline would naively match "stirring" or "peppers," retrieving incorrect videos. Our Reason-then-Retrieve framework succeeds through a two-stage process.

\noindent\textbf{Stage 1: After-Effect Reasoning.} The model generates a structured reasoning trace $R$ that explicitly predicts the visual consequences of the edit across multiple semantic dimensions: (i)~\emph{Actions}---the primary activity transitions from \texttt{chopping} to \texttt{stirring}; (ii)~\emph{Object States}---the \texttt{vegetables} become \texttt{diced} and are now contained in a \texttt{pan}; (iii)~\emph{Scene Context}---the setting shifts from a \texttt{cutting board} to a \texttt{stovetop}; and (iv)~\emph{Camera Viewpoint}---the instruction explicitly specifies a \texttt{close-up shot}.

\noindent\textbf{Stage 2: Effect-Aware Target Generation and Retrieval.} Conditioned on the reasoning trace $R$, the model generates a hypothetical target description that captures those predicted visual changes. The final query embedding $q(V_r, E)$ is derived from this effect-aware description (\emph{e.g.}, ``a close-up of stirring diced peppers in a pan on a stove''), enabling precise matching with the ground truth video while effectively rejecting distractors that lack these implied visual elements.

Figure~\ref{fig:success_reasoning_example} shows the detailed reasoning trace for the airplane landing example, illustrating how the model reasons about environmental changes, object dynamics, and camera perspective shifts.

\begin{figure}[t]
% \begin{tcolorbox}[colback=gray!5, colframe=gray!40, boxrule=0.5pt, arc=2pt]
\begin{tcolorbox}[colback=gray!5, colframe=gray!40, boxrule=0.5pt, arc=2pt, title={\small\textbf{Reasoning trace for the first example in Fig.~\ref{fig:success}}}]
\small
\textbf{Reasoning Trace:}

\vspace{2pt}
\noindent\textbf{1. Key elements affected:}
\begin{itemize}[leftmargin=15pt, itemsep=0pt, topsep=2pt]
   \item Sky background (currently dark night → twilight pinkish-orange)
   \item Plane's motion (currently blurred, distant approach → controlled descent, touchdown, deceleration)
   \item Visual effects (no smoke trail → faint smoke at touchdown)
   \item Environment (faint airport structures → distant trees and buildings)
   \item Perspective (plane remains large and centered → shrinks in distance as it decelerates)
\end{itemize}

\vspace{2pt}
\noindent\textbf{2. Specific transformations:}
\begin{itemize}[leftmargin=15pt, itemsep=0pt, topsep=2pt]
   \item The sky transitions from black to a soft pinkish-orange twilight, casting ambient warmth over the runway.
   \item The plane is reanimated to follow a descent arc, lowering its nose as it nears the runway, with touchdown marked by a faint smoke trail from tires.
   \item Background structures are replaced with silhouetted trees and distant urban buildings, enhancing contextual realism.
   \item Camera perspective adjusts to track the plane as it decelerates, smoothly pulling back to show its shrinking size while emphasizing the landing process.
\end{itemize}

\vspace{2pt}
\noindent\ldots \\
% \noindent\ldots
\end{tcolorbox}
% \caption{\textbf{Example reasoning trace.} for \ref{fig:success} first sample}
\caption{\textbf{Example reasoning trace for the first success case} (airplane landing in Figure~\ref{fig:success}). The trace explicitly predicts visual after-effects across multiple dimensions: environmental changes (sky, background), object dynamics (plane motion, smoke), and camera perspective—enabling precise target video matching.}
\label{fig:success_reasoning_example}
\vspace{-1.5em}
\end{figure}

\begin{figure}[t]
\begin{tcolorbox}[colback=gray!5, colframe=gray!40, boxrule=0.5pt, arc=2pt, title={\small\textbf{Reasoning trace for the second example in Fig.~\ref{fig:failure}}}]
\small
\textbf{Reasoning Trace:}

\vspace{2pt}
\noindent\textbf{1. Key elements affected:}
\begin{itemize}[leftmargin=15pt, itemsep=0pt, topsep=2pt]
   \item The rusted metal box (central subject) and padlock (on top) would be removed
   \item The metal fence, blue cylindrical object, and vegetation (grass, dirt) would be deleted
   \item The brick wall and overgrown background would be replaced by a beige building with arched architectural features
   \item The camera's initial pan right to the fence would be replaced by a focus on the street sign, followed by a shift to reveal a doorway
   \item No human presence in the source would be replaced with a person in a pink shirt entering from the right
\end{itemize}

\vspace{2pt}
\noindent\textbf{2. Specific transformations:}
\begin{itemize}[leftmargin=15pt, itemsep=0pt, topsep=2pt]
   \item The scene's visual tone shifts from decaying, neglected urban decay to a clean, maintained commercial facade
   \item A sign is added on to the street right in front of a beige building
   \item Architectural elements (arches, beige walls) replace the original rusted textures and overgrowth
   \item Camera movement changes from panning right to a focused reveal of the sign and then the doorway
   \item A person in a pink shirt is introduced, walking toward the building, adding narrative motion and human context
\end{itemize}

\vspace{2pt}
\noindent\ldots
\end{tcolorbox}
% \caption{\textbf{Reasoning trace example for the failure case} (pharmacy sign in Figure~\ref{fig:failure}). Despite capturing multiple transformation dimensions (objects, architecture, camera, human presence), the trace fails to preserve critical specificity (exact sign text), leading to semantic drift and retrieval failure with hyper-specific instructions.}
\caption{\textbf{Reasoning trace for the second failure case} (bottom one in Figure~\ref{fig:failure}). The trace captures broad transformations across multiple dimensions but loses critical details like exact sign text ("Crocker Pharmacy Palace"). This incomplete reasoning propagates through the pipeline, causing the generated target description to diverge from ground truth. This illustrates how hyper-specific instructions lead to compounding errors and semantic drift.}

\label{fig:reasoning_example}
\vspace{-1em}
\end{figure}

% \vspace{1em}
\subsection{Failure Case Analysis}

%Samples were Normal mode failed and the reasoning helped 

\begin{figure*}[t!]
    \centering
    \includegraphics[width=\textwidth]{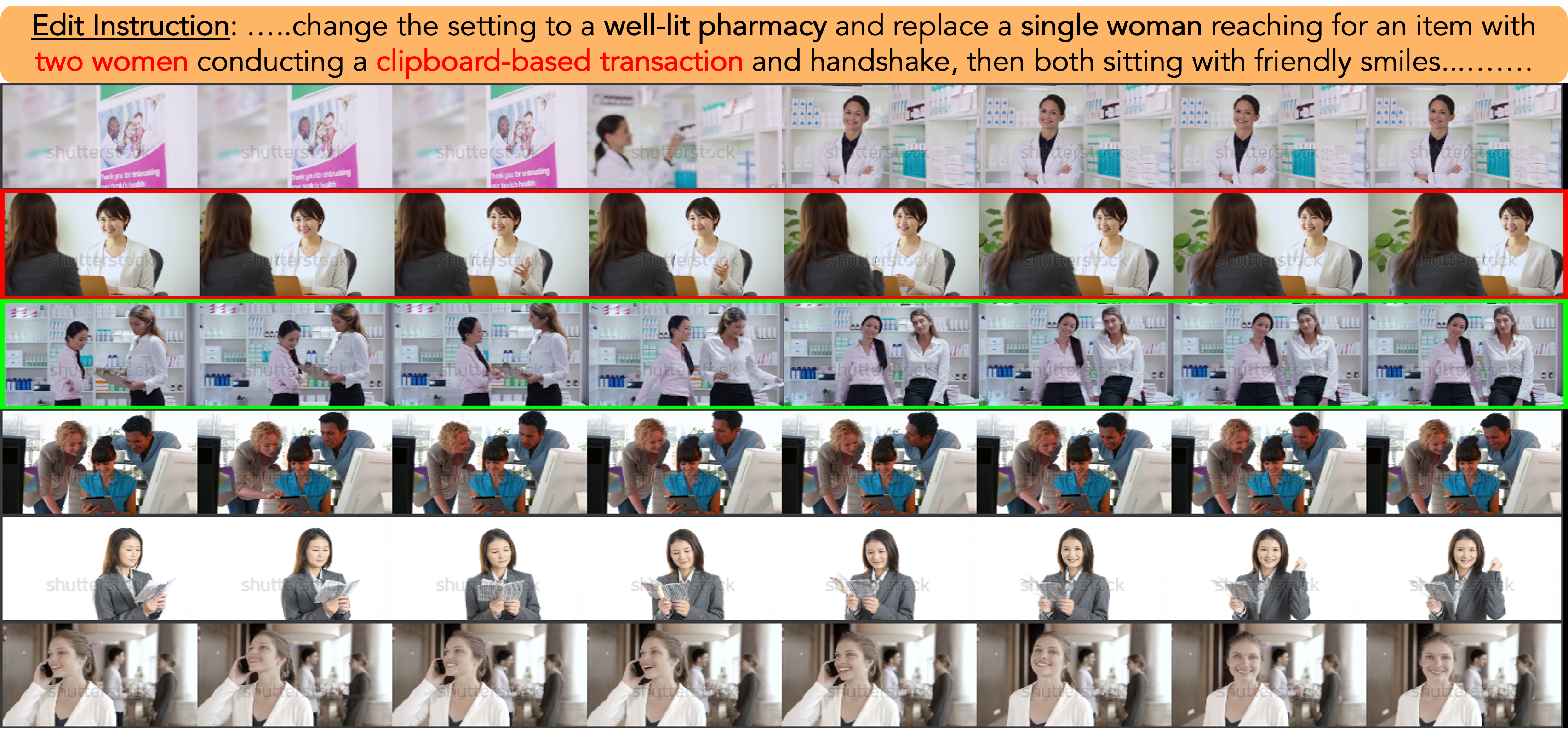}\\
    % \vspace{-1em}
    \includegraphics[width=\textwidth]{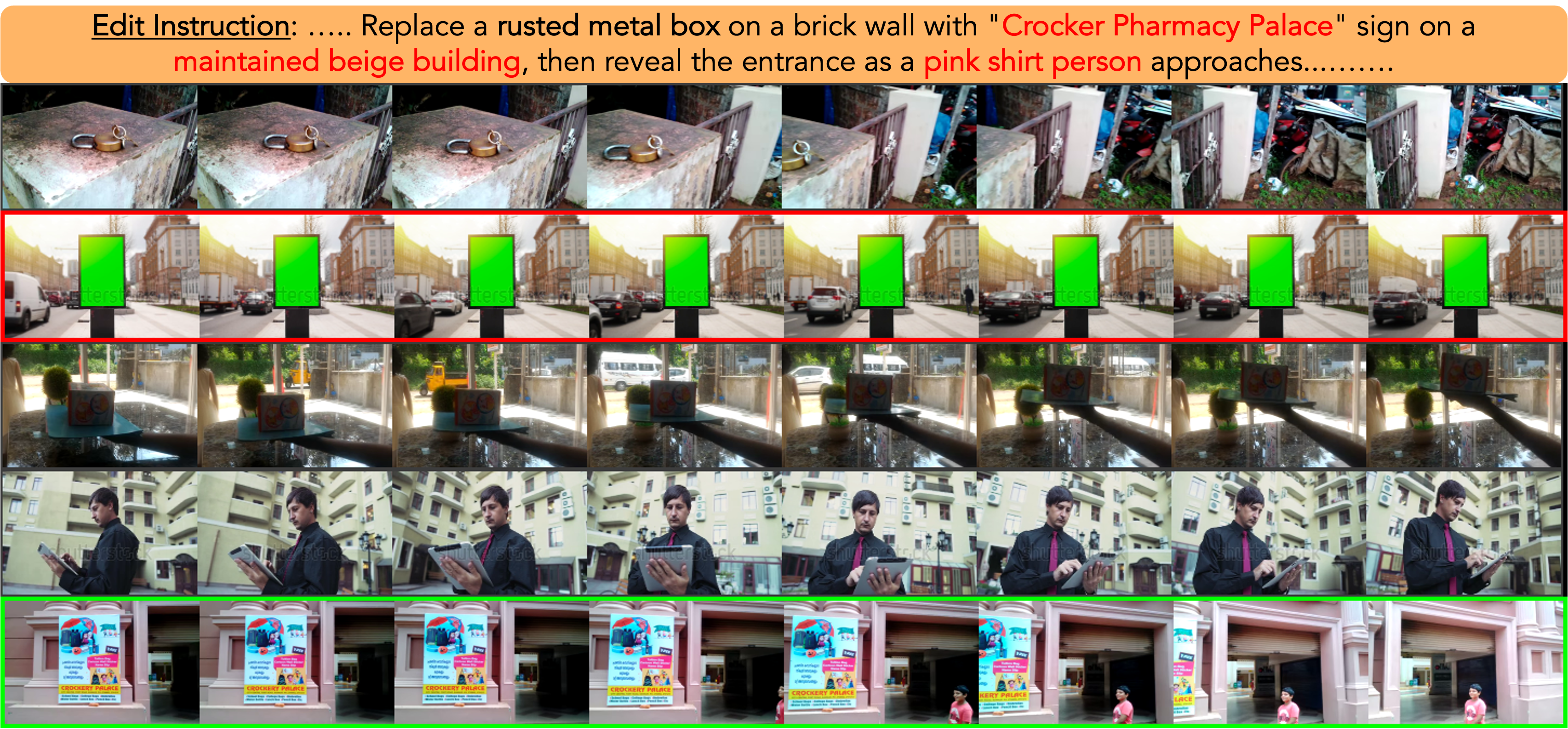}
    \caption{Two failure cases with incorrect Top-1 retrievals: two women smiling (top) and Crocker Pharmacy Palace sign (bottom). Each shows the edit instruction with key concepts highlighted in red, reference video (Row 1), and top-k predictions. Red borders indicate incorrect Top-1 retrievals; green borders indicate ground truth. Despite Top-1 errors, retrieved videos exhibit strong semantic alignment with instructions, demonstrating the model captures conceptual gist but loses the specificity needed for exact matching.}
    \label{fig:failure}
\end{figure*}

Figure~\ref{fig:failure} illustrates where our architecture encounters challenges: hyper-specific edit instructions with exhaustive detail. Consider the poster example (bottom), which requires replacing a ``rusted metal box'' with a ``Crocker Pharmacy Palace sign,'' changing a ``brick wall'' to a ``beige building with arched features,'' and introducing multiple other specific elements. Unlike the success cases involving coherent conceptual transformations, this instruction comprises a long list of disparate, fine-grained modifications.

Our zero-shot architecture fails on this example due to compounding errors across the reasoning-to-retrieval pipeline. In \textbf{Stage 1},  the LMM reasoner struggles to capture all distinct elements from this exhaustive instruction list, likely missing some fine-grained details such as the exact sign text, resulting in an incomplete reasoning trace $R$ as shown in ~\ref{fig:reasoning_example}. This degraded reasoning then propagates to \textbf{Stage 2}, where the generated target description $D_{\text{target}}$ no longer accurately reflects the ground truth video's visual content. Consequently, the final query embedding $q(V_r, E)$ represents a hypothetical video that slightly deviates from the actual target. Notably, the top-ranked retrieval \emph{does} contain a signboard with buildings—the model captures the \emph{semantic gist} (signage in urban context) but loses specificity, essentially generating a query for a semantically similar but ultimately \emph{different} video. 

% can end from a future direction perspective - diplomatic 
% \textcolor{red}{Examining the top-4 retrievals reveals strong semantic alignment: top-2 contain poster-like signage, ranks 2-3 include a person (matching the instruction), and rank 4 is ground truth. Even in failure cases, all retrievals exhibit visual similarity to the instruction's core concepts, reflecting the strong video understanding capabilities of large multimodal models. While parameter-efficient fine-tuning could potentially help capture distribution-specific nuances needed for exact top-1 matching, we leave this computationally expensive direction for future work.}

The top-4 retrieved results demonstrate strong semantic consistency: the first two images capture poster-like visual structures, the next two include a person as specified in the instruction, and the ground-truth appears within the top-4. Even when the model does not rank the exact match at the top, all retrieved samples remain semantically aligned with the key concepts. This behavior reflects the robust visual–semantic reasoning ability of large multimodal models. While parameter-efficient fine-tuning could further refine distribution-specific nuances required for precise top-1 retrieval, such exploration is computationally demanding and is left for future work.

% we have bigger claims -> its hard to justify things with a single example. 
% \textcolor{red}{This failure highlights a critical open question: What level of detail should an LMM capture in its intermediate representations to optimize downstream retrieval? Our analysis suggests that hyper-specific instructions can actually \emph{harm} performance by introducing noise that overwhelms the model's capacity to maintain coherent semantic understanding. When reasoning traces attempt to encode every granular detail, they may paradoxically become less effective retrieval signals than more abstracted, semantically-focused representations. Future work should investigate adaptive reasoning strategies that dynamically adjust description granularity based on instruction complexity, potentially filtering out less-discriminative details while preserving core semantic content. This points toward a broader challenge in vision-language systems: determining the optimal information bottleneck between detailed visual understanding and robust cross-modal retrieval.}
This failure case brings forward an important open question: How much detail should an LMM encode in its intermediate representations to best support downstream retrieval? Our observations indicate that overly detailed or overly literal instructions may unintentionally degrade retrieval accuracy. When the reasoning traces attempt to preserve every fine-grained aspect of the scene, they can introduce noise that hinders the model’s ability to maintain a coherent and discriminative semantic representation. In contrast, more abstracted and semantically focused descriptions often provide stronger retrieval signals.
These findings motivate future research on adaptive reasoning mechanisms that can modulate the granularity of generated descriptions based on the complexity of the instruction. Such mechanisms could selectively preserve discriminative visual cues while suppressing redundant or low-utility details. More broadly, this highlights a fundamental challenge for vision–language systems: identifying the optimal information bottleneck between detailed visual understanding and reliable cross-modal retrieval.

\section{Future Directions}

Our experiments reveal a fundamental tension: when does explicit reasoning help, and when does it just add overhead? This opens some promising directions.

\noindent\textit{Adaptive Routing.} WebVid-CoVR's simple edits (``add sunglasses'') favor keyword matching. CoVR-R's complex transformations demand causal reasoning. The solution? Route queries based on complexity: lightweight matching for simple edits, reasoning for complex ones. The challenge involves learning what makes an edit "complex." Is it length? Ambiguity? Implicit after-effects? Building robust complexity predictors remains open.

\noindent\textit{Learning Reasoning Depth.} Table~\ref{tab:reasoning_granularity} reveals something surprising: verbose traces hurt retrieval despite better reasoning scores. More isn't always better. Can we learn when to reason deeply versus stay abstract? Future work could explore controllers that start simple and extend thinking only when needed utilizing reinforcement learning to discover the optimal depth for each query. This would balance accuracy against cost, automatically determining when extended reasoning helps versus hurts.

\noindent\textit{Efficient Adaptation.} Our zero-shot approach generalizes well, yet Figure~\ref{fig:failure} shows an intriguing pattern: failures still capture semantic gist. The model understands but misses specifics. Lightweight techniques (eg: LoRA, adapters) could bridge this distribution gap without expensive retraining. The question: what minimal adaptation moves from gist to exact matching?

These directions share a theme: adaptive strategies that modulate reasoning based on query characteristics, moving beyond \emph{one-size-fits-all} approaches.

%% file: tables/webvid.tex
% \begin{table*}[t]
\centering
\setlength{\tabcolsep}{12pt}
\resizebox{\linewidth}{!}{%
\begin{tabular}{lcccc}
\toprule
\rowcolor{cyan!1} \textbf{Method} & \textbf{R@1} & \textbf{R@5} & \textbf{R@10} & \textbf{R@50} \\
\midrule
CoVR-BLIP \cite{webvid-covr} & 45.46 & 70.46 & 79.54 & 93.27 \\
CoVR-BLIP-2 \cite{webvid-covr} & 45.66 & 71.71 & 81.30 & 94.80 \\
BSE-CoVR \cite{thawakar2025beyond} & 48.08 & \textbf{73.36} & \textbf{81.06} & \textbf{93.78} \\
\midrule
\rowcolor{cyan!5} \textbf{Our Approach} & \textbf{49.15} & 70.72 & 79.25 & 93.42 \\
\bottomrule
\end{tabular}
}
\caption{\textbf{Comparison of our approach with existing methods on WebVid-CoVR test set}. Our approach achieves the highest R@1 (49.15\%), demonstrating strong performance on explicit modification instructions.}
% \end{table*}